\documentclass[acmsmall, review=false, anonymous=false, nonacm=true]{acmart}

\usepackage{verbatim}
\usepackage{mdframed}
\usepackage{amsmath}
\usepackage{xcolor}
\usepackage{bm} 
\usepackage{csquotes}
\usepackage{float}
\usepackage{subcaption}
\usepackage{wrapfig}
\usepackage{tikz}
\usepackage{listings}
\usepackage{graphbox}
\usepackage{xspace}
\usepackage{multicol}
\usepackage{mathtools}
\usepackage[most]{tcolorbox}
\usepackage{hyphenat}

\newcommand{\todo}[1]{\textcolor{black}{#1}}

\newcommand{\change}[1]{#1}
\newcommand{\edit}[1]{\textcolor{black}{#1}}
\newcommand{\edittwo}[1]{\textcolor{black}{#1}}

\newcommand{\V}{\textnormal{V}\xspace}
\newcommand{\vidx}[1]{\textit{v$_{#1}$}\xspace}
\newcommand{\vidxf}[1]{\textit{v$_{#1}^{\,F}$}\xspace}
\newcommand{\nbh}[1]{$\mathcal{N}_{#1}$\xspace}
\newcommand{\nbhf}[1]{$\mathcal{N}_{#1}^F$\xspace}
\newcommand{\numV}{\ensuremath{\lvert\V\rvert}\xspace}

\newcommand{\magline}[1]{\ensuremath{\lvert}\textnormal{#1}\ensuremath{\rvert}\xspace}

\newcommand{\E}{\textnormal{E}\xspace}
\newcommand{\eidx}[2]{\textit{e$_{#1#2}$}\xspace}
\newcommand{\eidxf}[2]{\textit{e$_{#1#2}^{\,F}$}\xspace}
\newcommand{\numE}{\ensuremath{\lvert\E\rvert}\xspace}

\newcommand{\hidx}[2]{\textit{h$_{#1}^{\,#2}$}\xspace}

\newcommand{\oidx}[1]{\textit{o$_{#1}$}\xspace}

\newcommand{\Gshort}{\textit{G}\xspace}
\newcommand{\Glong}{\textit{G}\,(\textit{\V, \E\hspace{-0.3mm}})\xspace}
\newcommand{\addth}[1]{\textit{$#1^{th}$}\xspace}

\newcommand{\fntrans}{${f}\:\xspace$}
\newcommand{\fnoutput}{${g}\:\xspace$}

\newcommand{\matdegree}{\textbf{D}\xspace}
\newcommand{\matadjacent}{\textbf{A}\xspace}
\newcommand{\matweight}{\textbf{A$_{W}$}\xspace}
\newcommand{\matlaplacian}{\textbf{L}\xspace}
\newcommand{\matidentity}{\textbf{I$_{n}$}\xspace}
\newcommand{\matincidence}{\textbf{M}\xspace}
\newcommand{\mateigenvec}{\textbf{U}\xspace}
\newcommand{\matlambda}{\bm{\Lambda}\xspace}

\newcommand{\real}{\hspace{0.5mm}\mathbb{R}}

\renewcommand*{\arraystretch}{.8}

\mdfdefinestyle{examplebox}{
    splittopskip=0,%
    splitbottomskip=0,%
    frametitleaboveskip=0,%
    frametitlebelowskip=0,
    skipabove=3cm,%
    skipbelow=3cm,%
    leftmargin=3cm,%
    rightmargin=3cm,
    innerleftmargin=1mm,%
    innerrightmargin=1mm,%
    innertopmargin=1mm,%
    innerbottommargin=1mm,
    roundcorner=0mm,
      linewidth=0.1pt,
      topline=true,
      bottomline=true,
      rightline=true,
      leftline=true}

\AtBeginDocument{%
  \providecommand\BibTeX{{%
    \normalfont B\kern-0.5em{\scshape i\kern-0.25em b}\kern-0.8em\TeX}}}

\setcopyright{acmlicensed}
\acmJournal{CSUR}
\acmYear{2021} \acmVolume{1} \acmNumber{1} \acmArticle{1} \acmMonth{1} \acmPrice{15.00}\acmDOI{10.1145/3503043}

\begin{document}

\title
[A Practical Tutorial on GNNs]
{A Practical Tutorial on Graph Neural Networks}



\subtitle{\edit{What are the fundamental motivations and mechanics that drive Graph Neural Networks, what are the different variants, and what are their applications?}}


\author{Isaac Ronald Ward$^{1}$}
\affiliation{%
  \institution{ISOLABS}
  \country{Australia}
}
\affiliation{%
  \institution{the University of Southern California}
  \country{USA}
}
\email{isaacronaldward@gmail.com}
\orcid{0000-0002-3418-3138}

\stepcounter{footnote}

\author{Jack Joyner$^1$}
\affiliation{%
  \institution{ISOLABS}
  \country{Australia}
}

\author{Casey Lickfold$^1$}
\affiliation{%
  \institution{ISOLABS}
  \country{Australia} 
}


\footnotetext{Equal contribution.}

\author{Yulan Guo}
\affiliation{%
  \institution{Sun Yat-sen University}
  \country{China}
}

\author{Mohammed Bennamoun}
\affiliation{%
  \institution{The University of Western Australia}
  \country{Australia}
}

\renewcommand{\shortauthors}{I. R. Ward, J. Joyner, C. Lickfold, S. Rowe, Y. Guo, M. Bennamoun}

\begin{abstract}
    

    
    Graph neural networks (GNNs) have recently grown in popularity in the field of artificial intelligence (AI) due to their unique ability to ingest relatively unstructured data types as input data. Although some elements of the GNN architecture are conceptually similar in operation to traditional neural networks (and neural network variants), other elements represent a departure from traditional deep learning techniques. This tutorial exposes the power and novelty of GNNs to AI practitioners by collating and presenting details regarding the motivations, concepts, mathematics, and applications of the most common and performant variants of GNNs. Importantly, we present this tutorial concisely, alongside practical examples, thus providing a practical and accessible tutorial on the topic of GNNs.
    
\end{abstract}



\maketitle





{\section{Introduction and Context}}
\label{s:intro}

Contemporary artificial intelligence (AI), or more specifically, deep learning (DL) has been dominated in recent years by the neural network (NN). NN variants have been designed to increase performance in certain problem domains; the convolutional neural network (CNN) excels in the context of image-based tasks, and the recurrent neural network (RNN) in the space of natural language processing (NLP) and time series analysis. NNs have also been leveraged as \edit{building blocks in more complex DL frameworks} --- for example, they have been used as trainable generators and discriminators in generative adversarial networks (GANs), and as \edit{components in Transformer networks} \cite{vaswani2017transformer}.

\begin{figure}[H]
    \centering
    
    \subfloat[A graph representation of a $14\times14$ pixel image of the digit `7'. Pixels are represented by vertices and their direct adjacency is represented by edge relationships.]{
        \includegraphics[width=0.45\textwidth]{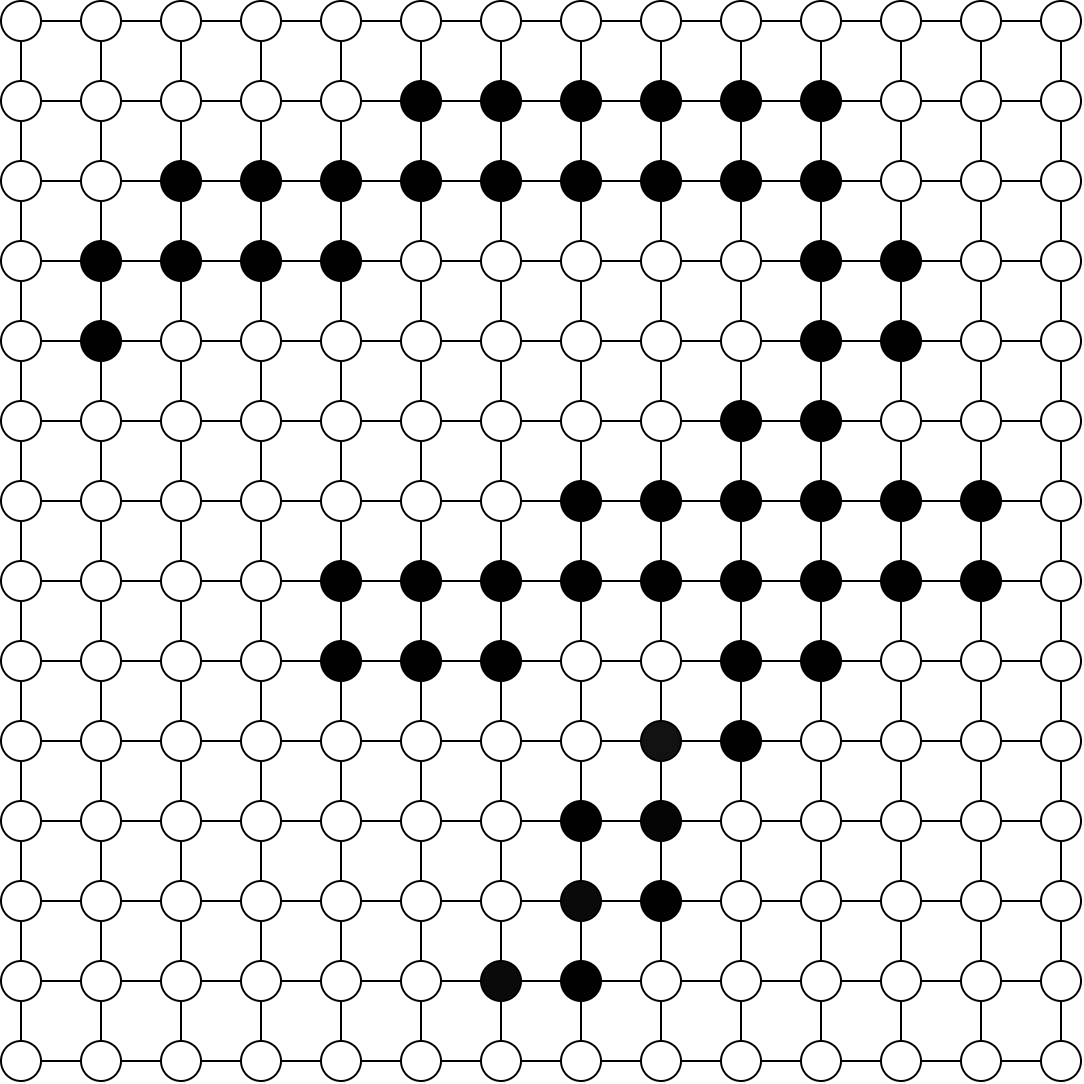}
    }\qquad
    \subfloat[\edit{A graph representing the joints in the human hand, and the hierarchical dependency of said joints. Images from the `Hands from Synthetic Data' dataset \cite{simon2017hands}.}]{
    \includegraphics[width=0.45\textwidth]{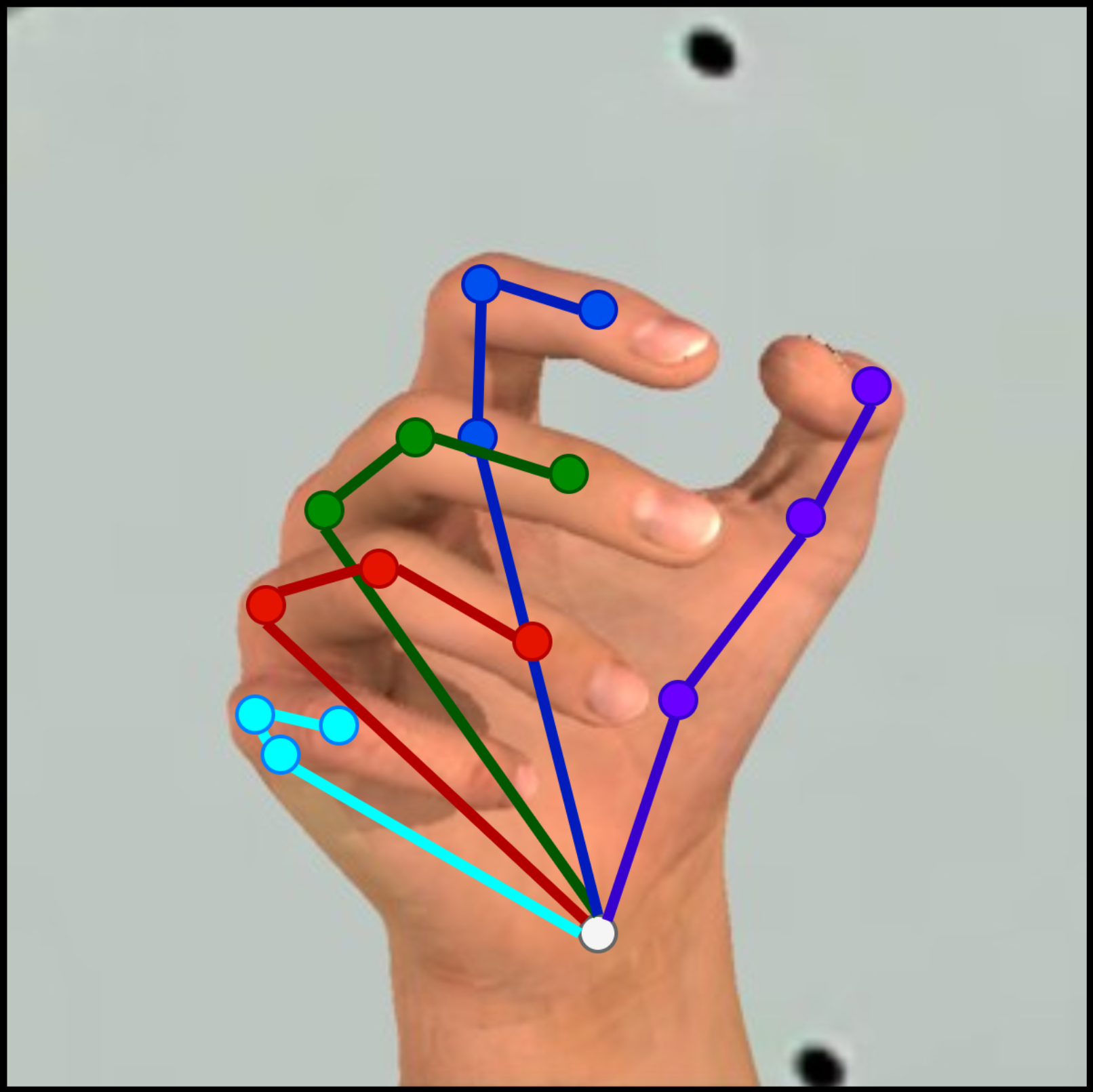}
    }
    \vspace{0.3cm}
    \\
    
    \subfloat[A diagram of an alcohol molecule (left), its associated graph representation with vertex indices labelled \\(middle), and its adjacency matrix (right).]{
    \begin{tabular}{c c p{5cm}}
        $\vcenter{\hbox{\includegraphics[width=0.32\textwidth]{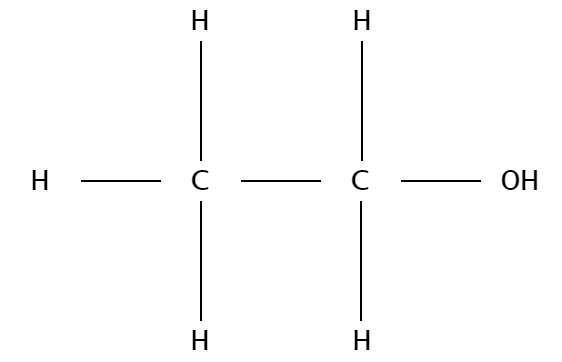}}}$ &
        $\vcenter{\hbox{\includegraphics[width=0.32\textwidth]{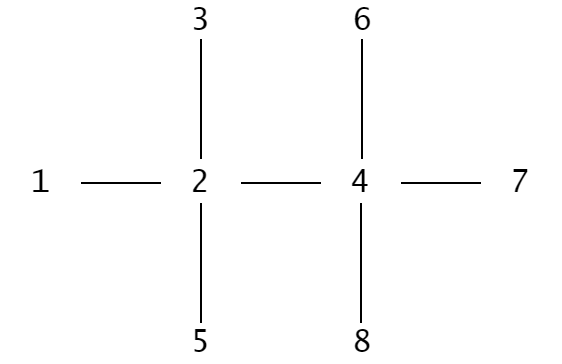}}}$ & 
        {$  \matadjacent = 
            \begin{bmatrix}
            0 & 1 & 0 & 0 & 0 & 0 & 0 & 0\\
            1 & 0 & 1 & 1 & 1 & 0 & 0 & 0\\
            0 & 1 & 0 & 0 & 0 & 0 & 0 & 0\\
            0 & 1 & 0 & 0 & 0 & 1 & 1 & 1\\
            0 & 1 & 0 & 0 & 0 & 0 & 0 & 0\\
            0 & 0 & 0 & 1 & 0 & 0 & 0 & 0\\
            0 & 0 & 0 & 1 & 0 & 0 & 0 & 0\\
            0 & 0 & 0 & 1 & 0 & 0 & 0 & 0\\
            \end{bmatrix}
        $}
    \end{tabular}
    }
    \vspace{5mm}
    \\ 
    
    \subfloat[A vector representation and a Reed--Kellogg diagram (rendered according to modern tree conventions) of the same sentence. The graph structure encodes dependencies and constituencies.]{
        \includegraphics[width=0.40\textwidth]{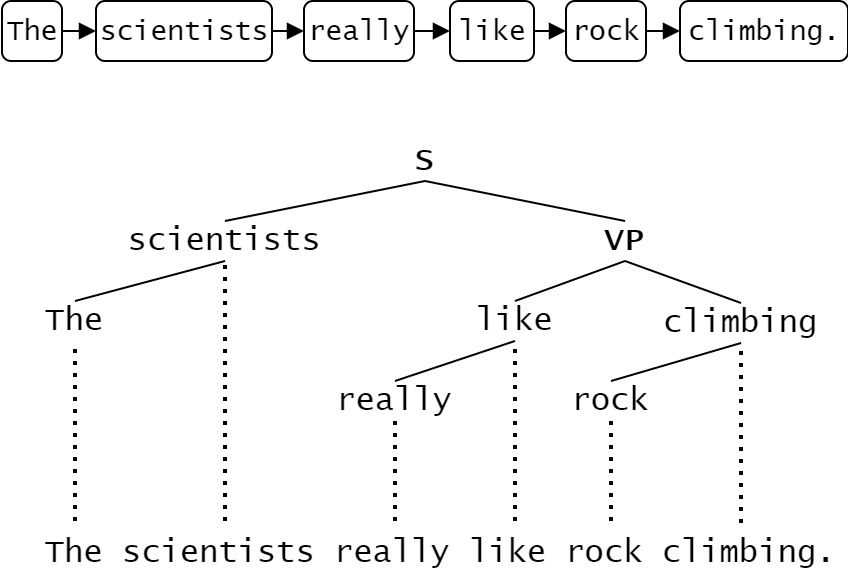}
    }\qquad
    \subfloat[A gameplaying tree can be represented as a graph. Vertices are states of the game and directed edges represent actions which take us from one state to another.]{
        \includegraphics[width=0.50\textwidth]{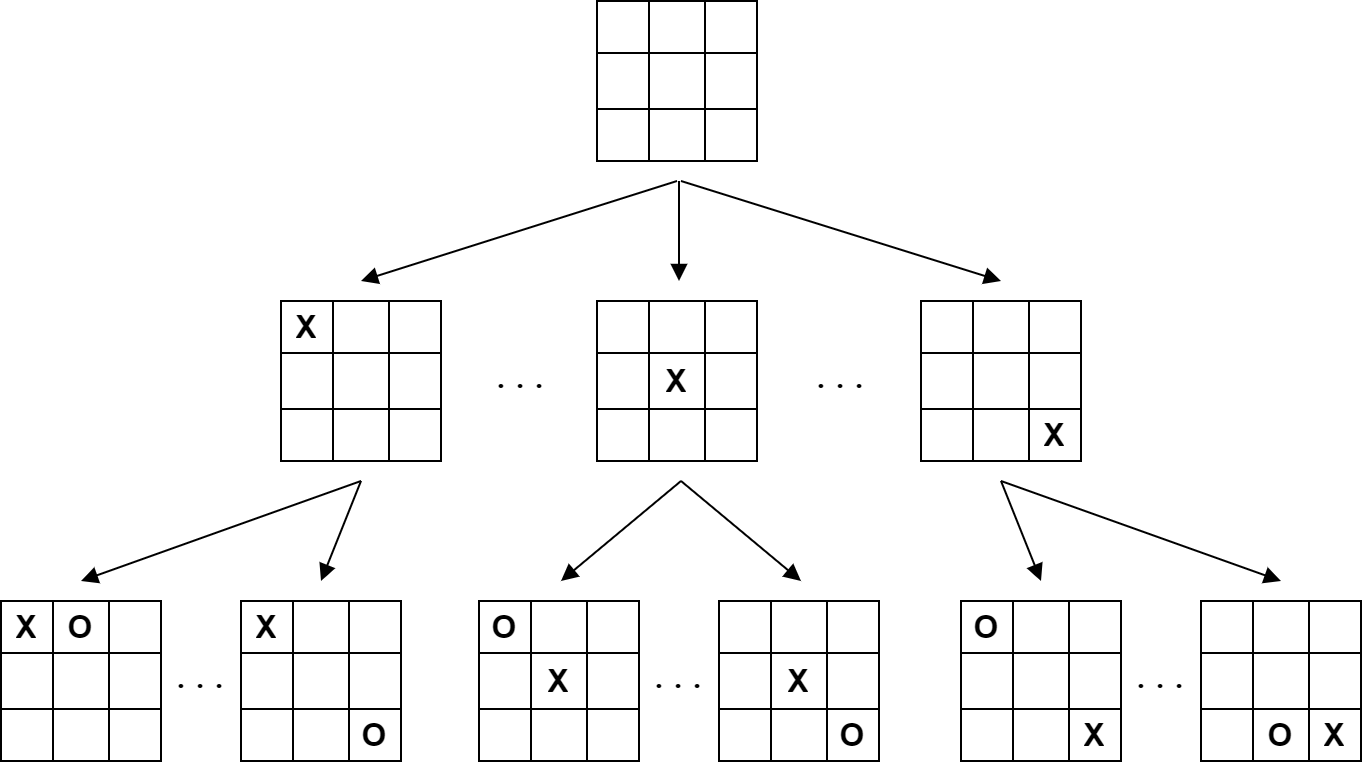}
    }
    \\
    
    \caption{The graphs data structure is highly abstract, and can be used to represent images (matrices), molecules, sentence structures, game playing trees, etc.}

    \label{fig:abstractgeneral}
\end{figure}

\edit{Graph neural networks (GNNs) provide a unified view of these input data types:} the images used as inputs in computer vision, and the sentences used as inputs in NLP can both be \edit{interpreted as special cases of} a single, general data structure --- \textbf{the graph} (see Figure~\ref{fig:abstractgeneral} for examples).

Formally, a graph is a set of distinct vertices (representing items or entities) that are joined optionally to each other by edges (representing relationships). Uniquely, the graphs fed into a GNN (during training and evaluation) \textbf{do not have strict structural requirements} per se; the number of vertices and edges between input graphs can change. In this way, GNNs can handle \textit{unstructured}, \textit{non-Euclidean} data \cite{bronstein2016geometric}, a property which makes them valuable in problem domains where graph data is abundant. Conversely, NN-based algorithms are typically required to operate on structured inputs with strictly defined dimensions. For example, a CNN built to classify over the MNIST dataset must have an input layer of $28 \times 28$ neurons, and all subsequent \edit{input images} must be $28 \times 28$ pixels in size to conform to this strict dimensionality requirement \cite{lecun2010mnist}. 

The expressiveness of graphs as a method for encoding data and the flexibility of GNNs with respect to unstructured inputs has motivated their research and development. They represent a new approach for exploring \edit{relatively} general \edit{DL} methods, and they facilitate the application of DL approaches to sets of data which --- until recently --- were not \edit{not exposed to AI}.

\subsection{Contributions}

\noindent{The \textit{key contributions} of this \edit{tutorial} paper are as follows}:

\begin{enumerate}
    \item \edit{An easy to understand, introductory tutorial, which assumes no prior knowledge of GNNs\footnote{\noindent We envisage that this work will serve as the `first port of call' for those looking to \edit{understand} GNNs, rather than as a comprehensive survey of methods and applications. For those seeking a more comprehensive treatment, we highly recommend the following works \cite{wu2019comprehensive, zhou2018methapps, zhang2018survey, hamilton2017methapps} (see Table~\ref{tab:contribnovel} for more detail).}.}
    \item Step-wise explanations of the \edit{mechanisms that underpin specific classes of GNNs, as enumerated in Table~\ref{tab:tax}.} These explanations progressively build a holistic understanding of GNNs.
    \item \edit{Descriptions of the advantages and disadvantages of GNNs, and key areas of application.}
    \item \edit{Full examples of how specific GNN variants can be applied to real world problems.}
\end{enumerate}















\edit{\subsection{Taxonomy}}

\noindent The structure and taxonomy of this paper is outlined in Table~\ref{tab:tax}.

\begingroup
\setlength\arraycolsep{2pt}
\renewcommand*{\arraystretch}{1.1}
\begin{table}[h]
    \begin{tabular}{p{4.5cm} p{8.4cm}}
        \toprule
        Broad class of algorithm & Related variants of algorithm \\
        \midrule
        
        Recurrent GNNs \newline(Section~\ref{s:rgnn}) & Graph LSTMs (Section~\ref{s:glstm}), \newline Gated GNNs (Section~\ref{s:gated}).   \\
        
        \midrule
        
        
        
        Convolutional GNNs  \newline(Section~\ref{s:cgnn}) & Spatial CGNNs (Section~\ref{s:cgnnspatial}, including Graph Attention Networks, Message Passing Neural Networks, etc.), \newline Spectral CGNNs (Section~\ref{s:cgnnspectral}). \\ 
        
        \midrule
        
        
        
        
        
        Graph Autoencoders \newline(Section~\ref{s:gaegeneral}) & Variational Graph Autoencoders (Section~\ref{s:vgae}), \newline Graph Adversarial Techniques (Section~\ref{s:gadvt}). \\
        
        \midrule
        
        
        
        

        \bottomrule
    \end{tabular}
\caption{A variety of algorithms are discussed in this tutorial paper. This table illustrates potential use cases for each algorithm, and the section where they are discussed. Should the reader prefer to read this tutorial paper from an \textit{applications} / \textit{downstream task-based} perspective, then we invite them to review Table~\ref{tab:rgnnapps}, Table~\ref{tab:cgnnapps}, and Table~\ref{tab:gaeapps}, which link each algorithm .  } 
\label{tab:tax}
\end{table}
\endgroup

\begingroup

\hyphenpenalty=10000
\exhyphenpenalty=10000

\setlength\arraycolsep{2pt}
\renewcommand*{\arraystretch}{1.1}
\begin{table}[H]
    \begin{tabular}{p{3.0cm} p{3.4cm} p{6.6cm}}
        \toprule
        GNN papers & Main sections & Description \\
        \midrule
        
        \vspace{-3.1mm}
        \begin{flushleft}
            This work 
        \end{flushleft} &
        
        
        \vspace{-3.1mm}
        \begin{flushleft}
        Recurrent GNNs, \newline
        Convolutional GNNs, \newline
        Graph Autoencoders \& \newline
        Graph Adversarial Methods 
        \end{flushleft} &
        
        \vspace{-3.1mm}
        \begin{flushleft}
        A \textbf{tutorial paper}  which steps through the operations of key GNN technologies in an explanatory and diagrammatic manner. Worked examples have been created to supplement explanations and are provided as code and in-text. 
        \end{flushleft} \\
        
        \midrule

        \vspace{-3.1mm}
        \begin{flushleft}
            Graph Neural Networks: A Review of Methods and Applications \cite{zhou2018methapps}
        \end{flushleft} &
        
        \vspace{-3.1mm}
        \begin{flushleft}
        GNN design framework, \newline
        GNN modules, \newline
        GNN variants, \newline
        Theoretical and \newline Empirical analyses \& \newline
        Applications 
        \end{flushleft} &
        
        \vspace{-3.1mm}
        \begin{flushleft}
        A \textbf{review paper} which proposes a general design framework for GNN models, and systematically elucidates, compares, and discusses the varying GNN modules which can exist within the components of said framework. 
        \end{flushleft}  \\
        
        \midrule

        \vspace{-3.1mm}
        \begin{flushleft}
            Deep Learning on Graphs: A Survey \cite{zhang2018survey}
        \end{flushleft} &
        
        \vspace{-3.1mm}
        \begin{flushleft}
        Recurrent GNNs, \newline
        Convolutional GNNs, \newline
        Graph Autoencoders, \newline
        Graph RL \& \newline
        Graph Adversarial Methods 
        \end{flushleft} &
        
        \vspace{-3.1mm}
        \begin{flushleft}
        A \textbf{survey paper} which outlines the development history and general operations of each major category of GNN. A complete survey of the GNN variants within said categories is provided (including links to implementations and discussions on computational complexity). 
        \end{flushleft} \\
        \midrule

        \vspace{-3.1mm}
        \begin{flushleft}
            A Comprehensive Survey on Graph Neural Networks \cite{wu2019comprehensive}
        \end{flushleft} &
        
        \vspace{-3.1mm}
        \begin{flushleft}
        Recurrent GNNs, \newline
        Convolutional GNNs, \newline
        Graph Autoencoders \& \newline
        Spatial-temporal GNNs 
        \end{flushleft} &
        
        \vspace{-3.1mm}
        \begin{flushleft}
        A \textbf{survey paper} which provides a comprehensive categorisation of contemporary GNN methods and benchmark datasets (across varying application domains). Numerous resources (e.g. open source code, datasets, etc.) are linked in a structured way. 
        \end{flushleft} \\

        \midrule

        \vspace{-3.1mm}
        \begin{flushleft}
            \edittwo{Computing graph neural networks: A survey from algorithms to accelerators \cite{abadal2021computing}}
        \end{flushleft} &
        
        \vspace{-3.1mm}
        \begin{flushleft}
        \edittwo{GNN fundamentals, \newline
        modeling, applications, \newline
        complexity, algorithms, \newline
        aceclerators \& data flows}
        \end{flushleft} &
        
        \vspace{-3.1mm}
        \begin{flushleft}
        \edittwo{A \textbf{review} of the field of GNNs is presented from a computing perspective. A brief \textbf{tutorial} is included on GNN fundamentals, alongside an \textbf{in-depth} analysis of acceleration schemes, culminating in a communication-centric vision of GNN accelerators. }
        \end{flushleft} \\
        
        \bottomrule
    \end{tabular}
\caption{A comparison of our tutorial and related works. While other works provide comprehensive overviews of the field, our work focuses on explaining and illustrating key GNN techniques to the AI practitioner. Our goal is to act as a `first port of call' for readers, providing them with a basic understanding that they can build upon when reading more advanced material.  }
\label{tab:contribnovel}
\end{table}

\endgroup

\section{Preliminaries}
\label{s:prelim}

\begingroup

\setlength\arraycolsep{2pt}
\renewcommand*{\arraystretch}{1.1}

\begin{table}[h]
  \begin{tabular}{l p{11cm}}
    \toprule
    Notation & Meaning \\
    \midrule
    
    \V                  & A set of vertices. \\
    \numV               & The number of vertices in a set of vertices \V. \\
    \vidx{i}            & The \addth{i} vertex in a set of vertices \V. \\
    \vidxf{i}           & The feature vector of vertex \vidx{i}. \\
    
    \E                  & A set of edges. \\
    \numE               & The number of edges in a set of edges \E. \\
    \eidx{i}{j}         & The edge between the \addth{i} vertex and the \addth{j} vertex, in a set of edges \E. \\
    \eidxf{i}{j}        & The feature vector of edge \eidx{i}{j}. \\
    
    \Gshort = \Glong    & A graph defined by the set of vertices \V and the set of edges \E. \\

    \nbh{\vidx{i}}      & The set of vertex \textit{indicies} for the vertices that are direct neighbors of \vidx{i}. \\
    
    \hidx{i}{k}         & The \addth{k} hidden layer's representation of the \addth{i} \edit{vertex. Since each layer typically aggregates information from neighbors 1-hop away, this representation includes information from neighbors k-hops away.}  \\   
    \oidx{i}            & The \addth{i} output of a GNN (indexing is \edit{dependant on output structure}). \\
    \matidentity        & An $n \times n$ identity matrix; all zero except for one's along the diagonal. \\
    
    \matadjacent        & The adjacency matrix; each element $\matadjacent_{ij}$ represents if the \addth{i} vertex is connected to the \addth{j} vertex by \edit{an edge}. \\
    
    \matdegree          & The degree matrix; a diagonal matrix of vertex degrees or valencies (the number of edges incident to a vertex). Formally defined as $\matdegree_{i,i} = \sum_{j} \matadjacent_{ij}$. \\
    
    \matweight        & The weight matrix; each element $\matweight_{ij}$ represents the `weight' of the edge between the \addth{i} vertex and the \addth{j} vertex. The `weight' typically represents some real concept or property. For example, the weight between two given vertices could be inversely proportional to their distance from one another (i.e., close vertices have a higher weight between them). Graphs with a weight matrix are referred to as \textit{weighted graphs}, but not all graphs are weighted graphs; \edit{in unweighted graphs \matweight $=$ \matadjacent. } \\
    
    \matincidence       & The incidence matrix; a \numV $\times$ \numE matrix where for each edge \eidx{i}{j}, the element of $\matincidence$ at $(i, \eidx{i}{j}) = +1$, and at $(j, \eidx{i}{j}) = -1$. All other elements are set to zero. \matincidence describes the incidence of all edges to all vertices in a graph. \\
    
    \matlaplacian       & The non-normalized \edit{combinatorial} graph Laplacian; defined as $\matlaplacian = \matdegree - \matweight$. \\
    
    \matlaplacian$\!_{sn}$       & The symmetric normalized graph Laplacian; defined as $\matlaplacian = \matidentity - \matdegree^{-\frac{1}{2}} \matadjacent \matdegree^{-\frac{1}{2}}$. \\
    
    
  \bottomrule
\end{tabular}
  \caption{Notation used in this work. We suggest that the reader familiarise themselves with this notation before proceeding.}
  \label{tab:not}
\end{table}

\endgroup




Here we discuss some basic elements of graph theory, as well as the the key concepts required to understand how GNNs are formulated and operate. We present the notation which will be used consistently in this work (see Table~\ref{tab:not}).

\subsection{Key Terms}
\label{s:defs}

\textbf{Graphs} are formally defined by a set of vertices and the set of edges between these vertices: put formally, \Gshort = \Glong. Fundamentally, graphs are just a way to encode data, and in that way, every property of a graph represents some real element, or concept in the data. Understanding how graphs can be used to represent complex concepts is key in appreciating their expressiveness and generality as an encoding device (see Figures~\ref{fig:abstractgeneral} for examples of this domain agnostic expressiveness).

\textbf{Vertices} represent items, entities, or objects, which can naturally be described by quantifiable attributes and their relationships to other items, entities, or objects. We refer to a set of \numV vertices as \V and the \addth{i} single vertex in the set as \vidx{i}. Note that there is no requirement for all vertices to be homogenous in their construction.


\textbf{Edges} represent and characterize the relationships that exist between items, entities, or objects. Formally, a single edge can be defined with respect to two (not necessarily unique) vertices. We refer to a set of \numE edges as \E and a single edge between the \addth{i} and \addth{j} vertices as \eidx{i}{j}. 


\textbf{Neighborhoods} are \textit{subgraphs} within a graph, \edit{and} represent distinct groups of vertices and edges. \edit{Most commonly, the neighborhood \nbh{\vidx{i}} centrerd around a vertex \vidx{i} comprises of \vidx{i}, its adjoining edges (where \eidx{i}{j} $= 1$), and the vertices that are directly connected to it.} Neighborhoods can be iteratively grown from a single vertex by considering the vertices attached (via edges) to the current neighborhood. Note that a neighborhood can be defined subject to certain vertex and edge feature criteria \edit{(i.e., all vertices within $2$ hops of the central vertex, rather than $1$ hop)}. 

\textbf{Features} are quantifiable attributes which characterize a phenomenon that is under study. In the graph domain, features can be used to further characterize vertices and edges. Extending our social network example, we might have features for each person (vertex) which quantifies the person's age, popularity, and social media usage. Similarly, we might have a feature for each relationship (edge) which quantifies how well two people know each other, or the type of relationship they have (familial, colleague, etc.). In practice there might be many different features to consider for each vertex and edge, so they are represented by numeric feature vectors referred to as \vidxf{i} and \eidxf{i}{j} respectively.




\textbf{Embeddings} are compressed feature representations. If we reduce large feature vectors associated with vertices and edges into low dimensional embeddings, it becomes possible to classify them with low-order models (i.e., if we can make a dataset linearly separable). A key measure of an embedding's quality is if the points in the original space retain the same similarity in the embedding space. Embeddings can be created (or learned) for vertices, edges, neighborhoods, or graphs. Embeddings are also referred to as representations, encodings, latent vectors, or high-level feature vectors depending on the context.

\edit{
\textbf{Output types} change depending on the problem domain.
A GNN's forward pass can be thought of as two key processes: converting input graphs into useful embeddings, performing some downstream task (e.g. classification) on the embeddings, which converts the embeddings into some useful output. We define three commonly observed output types as follows.
\begin{enumerate}
    \item \textbf{Vertex-level} outputs require a prediction (e.g. a distinct class or regressed value) for each vertex in a given graph.
    \item \textbf{Edge-level} outputs require a prediction for each edge in a given graph.
    \item \textbf{Graph-level} outputs require a prediction per graph. For example: predicting the properties molecule graphs \cite{wieder2020compact}.
\end{enumerate}
}

\edit{
\subsection{Learning Types}
\textbf{Transductive learning} methods are exposed to all of the training and testing data before making predictions. For example: our dataset might consist of a \textit{single large graph} (e.g. Facebook's social network graph) and the set of vertices is only partially labelled. The training set consists of the labelled vertices, and the testing set consists of both a small set of labelled vertices (for benchmarking) and the remaining unlabelled vertices. In this case, our learning methods should be exposed to the entire graph during training (including the test vertices), because the additional information (e.g. structural patterns) will be useful to learn from. Transductive learning methods are useful in such cases where it is challenging to separate the training and testing data without introducing biases.
\\\textbf{Inductive learning} methods reserve separate training and testing datasets. The learning process ingests the training data, and then the learned model is tested using the testing data, which it has not observed before in any capacity. 
}



\section{Recurrent Graph Neural Networks}
\label{s:rgnn}

In a standard NN, successive layers of learned weights work to extract \edit{progressively higher level} features from an input \edit{tensor}. In the case of \edit{NNs for computer vision}, the presence of low-level features --- such as short lines and curves --- are identified by earlier layers, whereas the presence of high-level features \edit{--- such as composite shapes ---} are identified by later layers. \change{After being processed by \edit{these} sequential layers, the resultant high-level features can then be provided to a softmax layer or single neuron for the purpose of classification, regression\edit{, or some other downstream task}. }

In the same way, \edit{the earliest GNNs extracted high-level feature representations from graphs by using successive feature extraction operations \cite{scarselli2009gnn, micheli2009neural}, and then routed these high-level features to output functions. In other words: they processed inputs into useful embeddings and then processed embeddings into useful outputs using two distinct stages of processing. These early techniques had limitations: some algorithms could only process Directed Acyclic Graphs (DAGs) \cite{di2006comparison}, others required the input graphs to have `supersource' vertices (which had directed paths to all other vertices in the graph) \cite{bianchini2002recursive}, and some techniques required heuristic approaches to deal with the cyclical nature of certain graphs \cite{micheli2001analysis}. }

\edit{Typically, these early \textit{recursive} methods relied on `unfolding' special cases of graphs into finite trees (recursive equivalents), which could then be processed into useful embeddings by recursive NNs \cite{bianchini2002recursive}. The \textit{Recurrent} GNN extended this, and thus provided a solution which could be applied to generic graphs \cite{scarselli2005graph}. Rather than create an embedding for the whole input graph via a recursive encoding network, RGNNs create embeddings at the vertex-level through an information propagation framework known as \textit{message passing}, which will be defined in this section.}








\subsection{Recurrently Computing Embeddings}

\begin{figure}[H]
    \centering
    
    \includegraphics[width=0.85\textwidth]{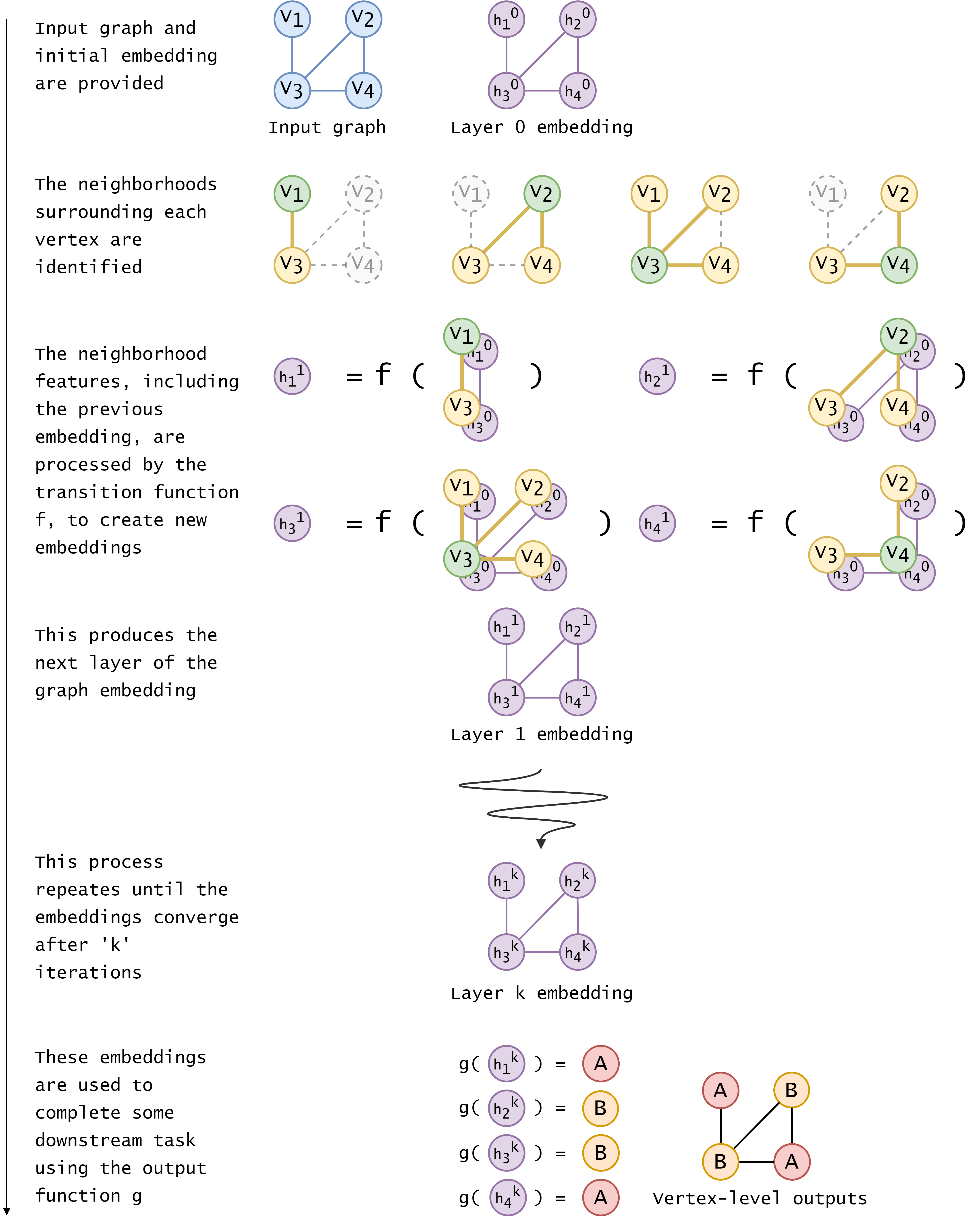}
    
    \caption{An RGNN forward pass for a simple input graph \Glong with \numV$=4$, \numE$=4$. \Gshort goes through $k$ layers of processing. In each layer, each vertex's features \vidxf{i} (green), the neighborhood's features \nbhf{\vidx{i}} (yellow), and the previous hidden layer (purple) are processed by the \textit{state transition function} \fntrans and aggregated, thereby producing successive embeddings of \Gshort. Note that the neighborhood features must be aggregated into a fixed embedding size, otherwise \fntrans would need to handle variable input sizes. This is repeated until the embeddings converge (i.e., the change between consecutive embeddings fails to exceed some stopping threshold). At that stage, the embeddings are fed to an \textit{output function} \fnoutput which perform some downstream task --- in this case, the task is a vertex-level classification problem. Note that \fntrans and \fnoutput can be implemented as NNs and trained via backpropagation of supervised error signals through the unrolled computation graph \cite{scarselli2009gnn, micheli2009neural}. Note that each vertex's embedding includes information from at max $k$ `hops' away after the \addth{k} layer of processing. Image best viewed in colour. }
    
    \label{fig:rec}
\end{figure}

RGNNs compute embeddings at each vertex in the input graph using a deterministic, shared function called the \textit{transition function}. \edit{It is named the transition function as it can be interpreted as calculating the \textit{next representation} of a neighborhood from the neighborhood's \textit{current representation}. This transition function can be applied symmetrically at any vertex, even though the size of a vertex's neighborhood may be variable. This process is illustrated in Figure~\ref{fig:rec}, where the transition function \fntrans calculates an embedding at each vertex, for the surrounding neighborhood.}

As such, the \addth{k} embedding \hidx{i}{k} for any given vertex \vidx{i} is dependent on the following quantities:

\begin{itemize}
    \item The features of the central vertex \vidxf{i}.
    \item The features of all adjoining edges \eidxf{i}{j}, $j\in$\nbh{\vidx{i}} (if edge features are present).
    \item The features of all neighboring vertices \vidxf{j}, $j\in$\nbh{\vidx{i}}.
    \item The previous iteration's embeddings of all neighboring vertices' \hidx{j}{k-1}, $j\in$\nbh{\vidx{i}}. \edit{\hidx{i}{0} $\forall i \in$ \V can be defined defined arbitrarily on initialisation, and Banach's fixed point theorem will guarantee that the subsequently calculated embeddings will converge to some optimal value exponentially (if \fntrans is implemented as a contraction map) \cite{khamsi2001banach}.}
\end{itemize}


In order to recurrently apply this learned transition function to compute successive embeddings, \fntrans must have a fixed number of input and output variables. How then can it be dependent on the immediate neighborhood, which might vary in size depending on where we are in the graph? \edit{There are two simple solutions, the first of which is to set a `maximum neighborhood size' and use null vectors when dealing with vertices that have non-existing neighbors \cite{scarselli2009gnn}. The second approach is to aggregate all neighborhood features in some permutation invariant manner \cite{gilmer2017neural}, thus ensuring that any neighborhood in the graph is represented by a fixed size feature vector. While both approaches are viable, the first approach does not scale well to `scale-free graphs', which have degree distributions that follow a power law. Since many real world graphs (e.g. social networks) are scale-free \cite{sanders2016scalable}, we'll use the second solution here. Mathematically, this can be formulated as in Equation~\ref{eq:rgnn} \cite{scarselli2009gnn}.}


\begin{equation}
    \hidx{i}{k} = \sum_{j\in\textnormal{\nbh{\vidx{i}}}}  \textnormal{\fntrans} (\vidxf{i}, \eidxf{i}{j}\,, \vidxf{j},  \hidx{j}{k-1}),\;\textnormal{where all \hidx{i}{0} are defined on initialisation.}
    \label{eq:rgnn}
\end{equation}


We can see that under this formulation \edit{Equation~\ref{eq:rgnn}}, \fntrans is well defined. It accepts four feature vectors which all have a defined length, regardless of which vertex in the graph is being considered, regardless of the iteration. This means that the transition function can be applied \edit{iteratively}, until a stable \edit{embedding} is reached for all vertices in the input graph. This expression can be interpreted as passing `messages', or features, throughout the graph; in every iteration, the embedding \hidx{i}{k} is dependant on the features and embeddings of its neighbors. This means that with enough recurrent iterations, information will propagate throughout the whole graph: after the first iteration, any vertex's embedding encodes the features of the neighborhood within a range of a single edge.

In the second iteration, any vertex's embedding is an encoding of the features of the neighborhood within a range of two edges away, and so on. The iterative passing of `messages' to generate an encoding of the graph is what gives this \textit{message passing} framework its name\footnote{\edit{Importantly, this is not the formulation Message Passing Neural Network (MPNN) model \cite{gilmer2017neural}, rather, it is a technique which uses the message passing framework}. \edit{State-of-the-art approaches will be discussed in Section~\ref{s:cgnnspatial}, and in particular, the MPNN will be defined explicitly.}}. 

\edit{Note that it is typical to explicitly add the identity matrix \matidentity to the adjacency matrix \matadjacent, thus ensuring that all vertices become trivially connected to themselves, meaning that a vertex \vidx{i} $\in$ \nbh{\vidx{i}} $\forall i \in \V$. Moreover, this allows us to directly access the neighborhood by iterating through a single row of the adjacency matrix. This modified adjacency matrix is usually normalised to prevent unwanted scaling of embeddings.}

\subsection{Computing downstream outputs}







Once we have useful embeddings centred around each vertex in the graph, the goal is to then inference meaningful outputs based on these values (i.e., to perform a \textit{downstream} task). The \textbf{output} function \fnoutput is responsible for taking the converged embeddings of a graph \Glong and creating said output. In practice, the output function \fnoutput, much like the transition function \fntrans, is implemented by a feed-forward neural network, though other means of returning a single value have been used, including mean operations, dummy super nodes, and attention sums \cite{zhou2018methapps}. 

\edit{Intuitively, the combined process of recurrently computing embeddings and subsequently computing downstream inputs can be interpreted as a sequential process of repeated NN computation blocks --- or a finite computation graph (see Figure~\ref{fig:rec}). In a supervised setting, a loss signal can be calculated which quantifies the error between the predicted output and a labelled ground truth. Both \fntrans and \fnoutput can then be trained via backpropagation of errors, throughout the `unrolled' computation graph. For more detail on this process, see the calculations in \cite{scarselli2009gnn}.}




\edit{\subsection{Extensions for Sequential Graph Data}}

When discussing \textit{recurrence} thus far, we have referred mainly to computing techniques that are iteratively applied to neighborhoods in a graph to produce embeddings that are dependent on information propagated throughout the graph. However, recurrent techniques may also refer to computing processes over \textit{sequential data}, e.g., time series data. In the graph domain, sequential data refers to instances which can be interpreted as graphs with features that change over time. These include spatiotemporal graphs \cite{wu2019comprehensive}. For example Figure~\ref{fig:abstractgeneral} (b) illustrates how a graph can represent a skeletal structure in a single image of a hand, however, if we were to create such a graph for every frame of a contiguous video of a moving hand, we would have a data structure that could be interpreted as a sequence of individual graphs, or a \textit{single} graph with sequential features, and such data could be used for classifying hand actions in video. 

\begin{figure}[H]
    \centering
    
    \subfloat[The general processing approach for sequential graph data. The input data is a sequence of graphs (blue), and each processing cell considers not only the current input state, but also information from the states preceding it, thus yielding per-state embeddings (purple) which are dependent on the sequence thus far. ]{
        \includegraphics[width=0.45\textwidth]{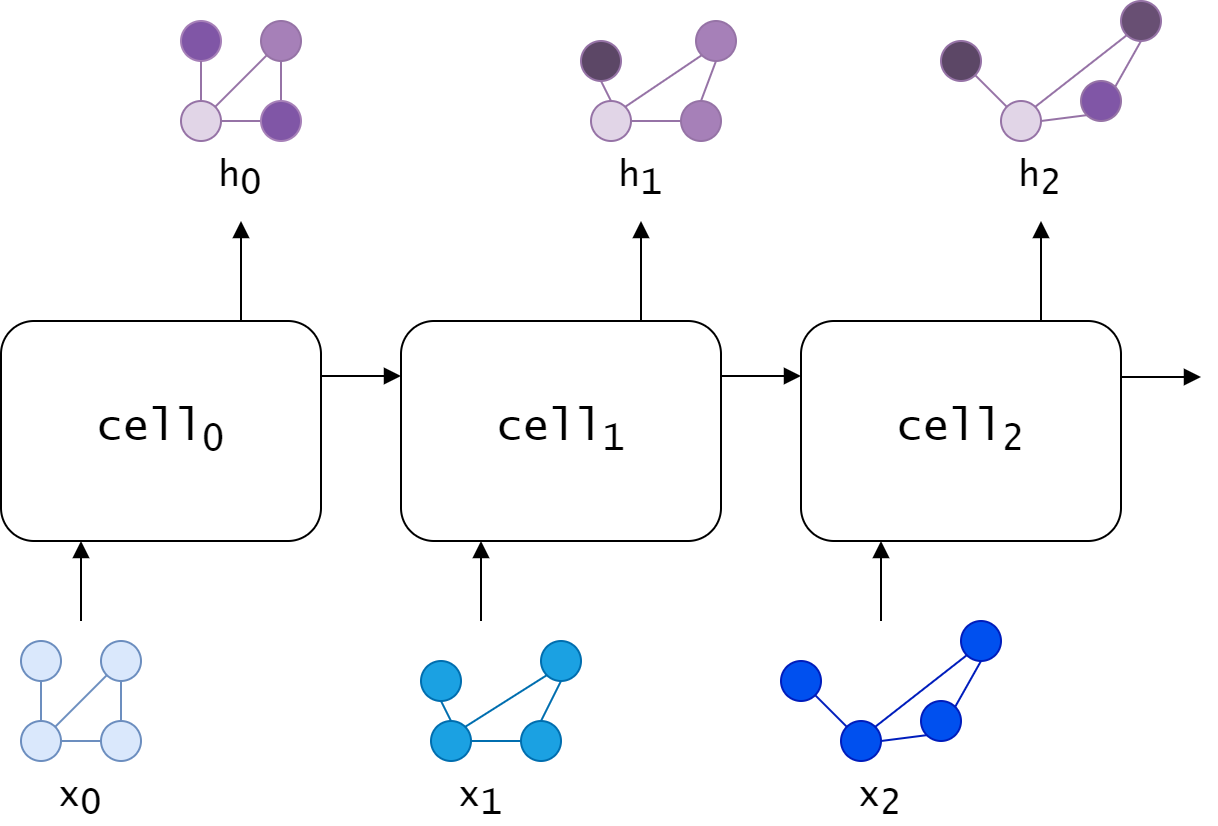}
    }\qquad
    \subfloat[A simple recurrent cell, which learns to extract useful features from the current input and the previous hidden state. After numerous cells of this type, the signal from the early input states is exponentially reduced. ]{
        \includegraphics[width=0.45\textwidth]{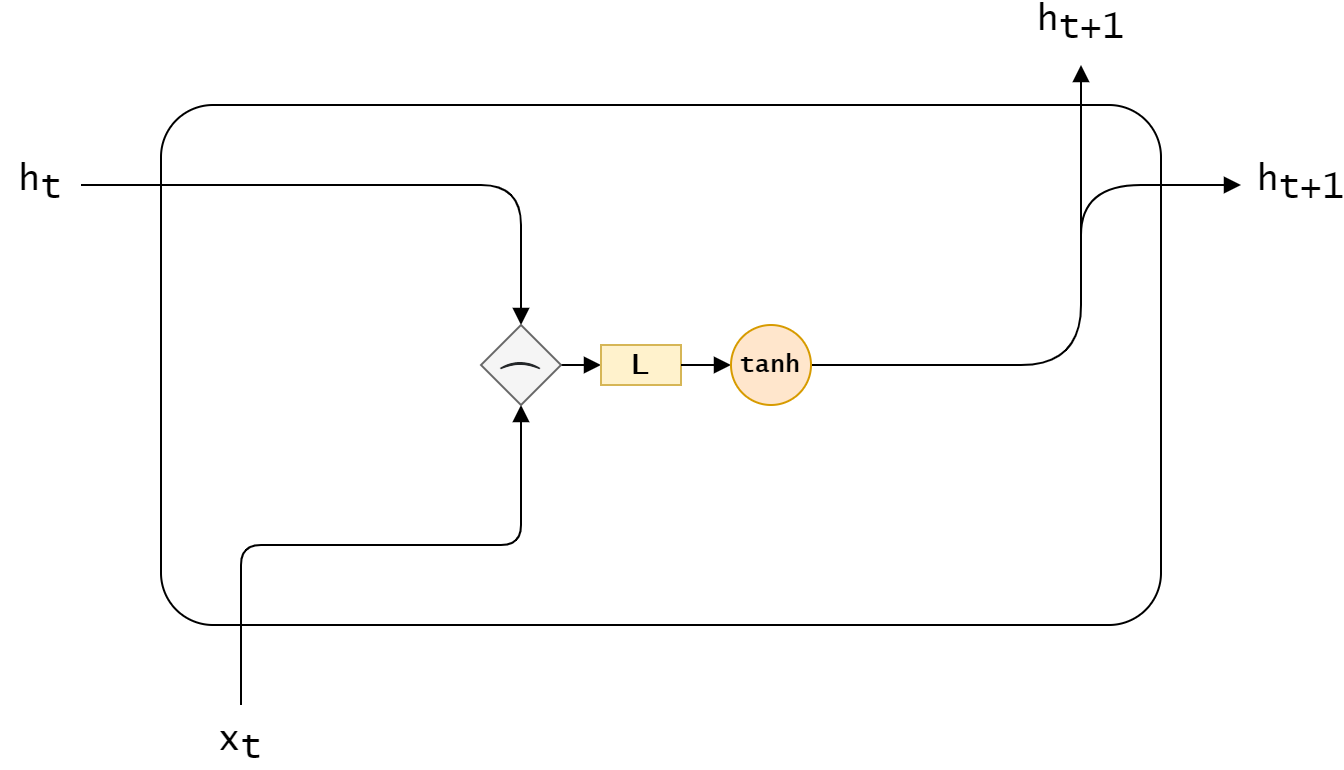}
    }
    \vspace{0cm}
    \\
    
    \subfloat[A GLSTM, which employs graph concatenation to enable LSTM-like processing of inputs. \textbf{Four} learned gates are employed to learn specific tasks within the cell. ]{
        \includegraphics[width=0.45\textwidth]{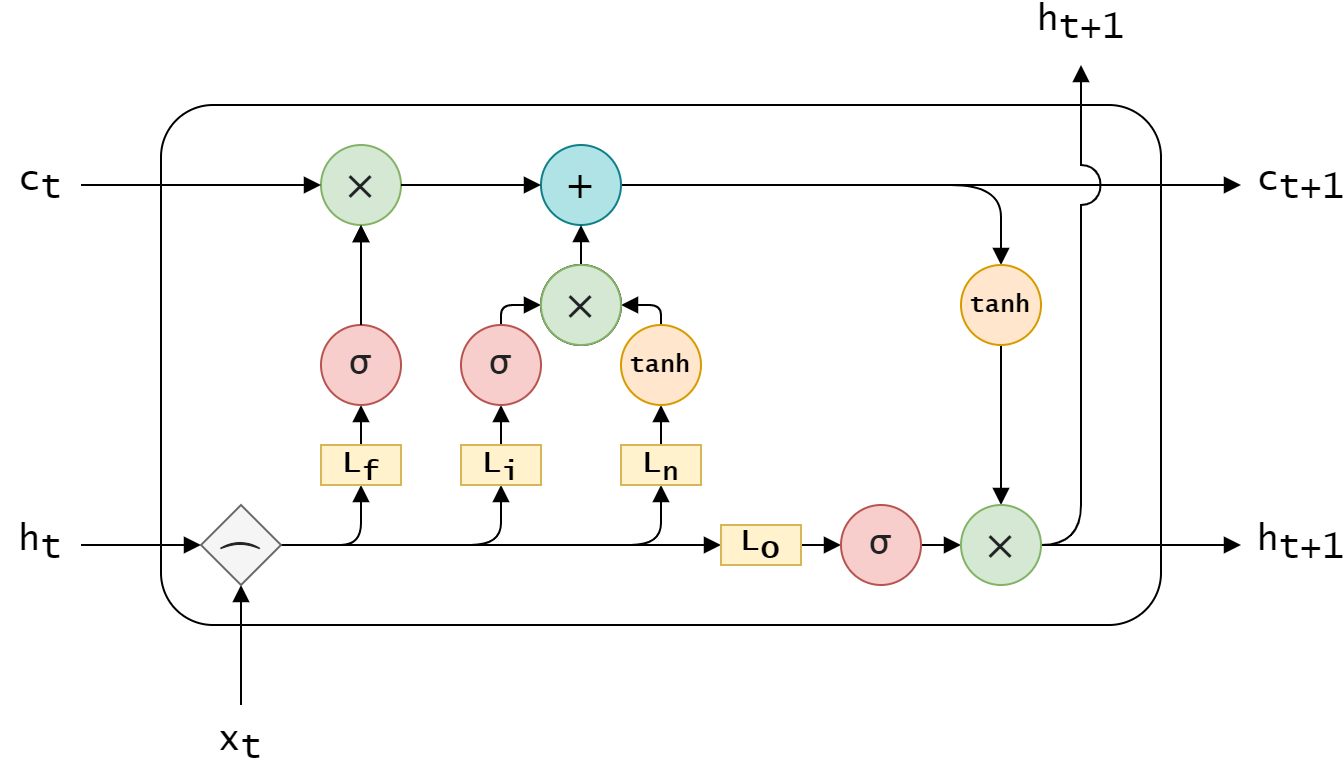}
    }\qquad
    \subfloat[A Graph GRU, which employs graph concatenation to enable GRU-like processing of inputs. GRUs have relatively less learnable parameters than LSTMs, and are generally less prone to overfitting. ]{
        \vspace{-0.9mm}
        \includegraphics[width=0.45\textwidth]{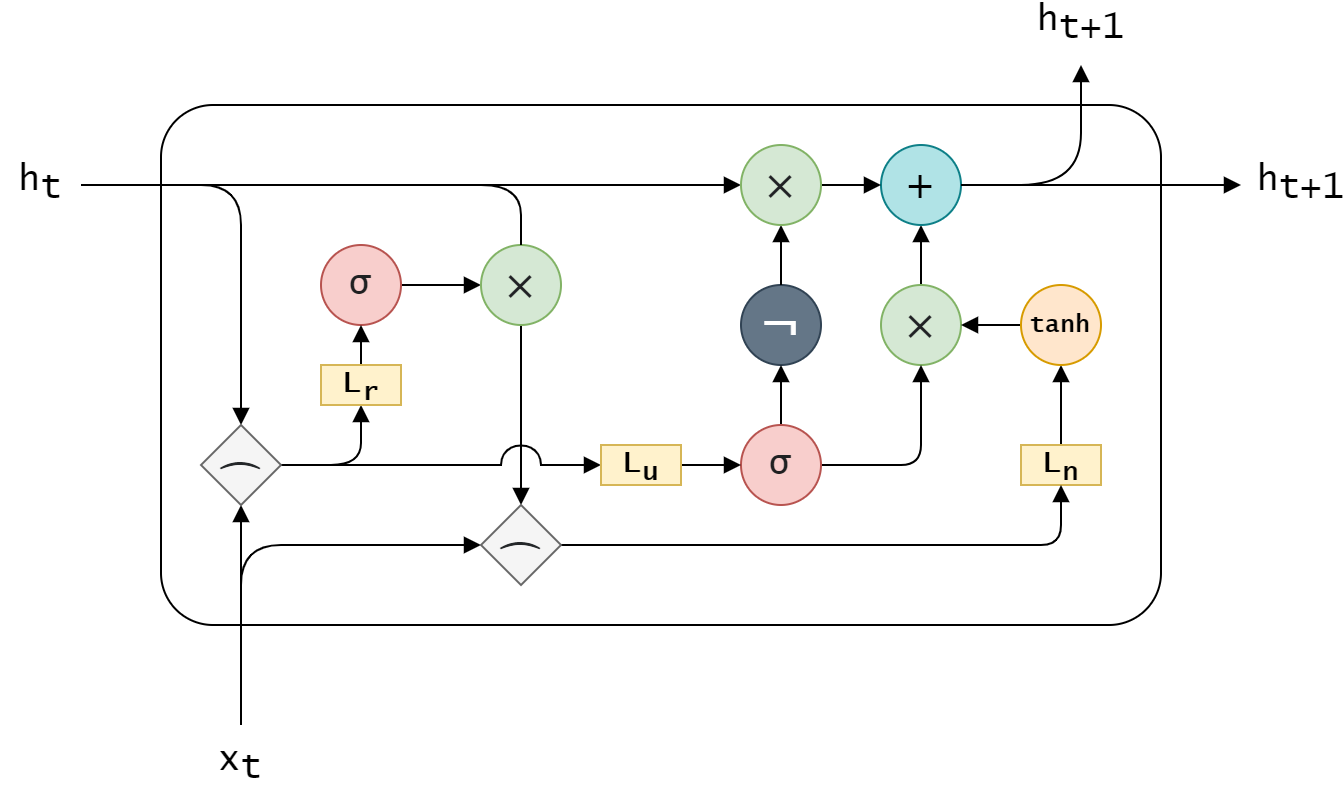}
    }
    \vspace{0cm}
    \\

    \subfloat[Legend for the diagrams (a) -- (d), all operators are traditional tensor operations apart from graph concatenation. ]{
        \includegraphics[width=0.8\textwidth]{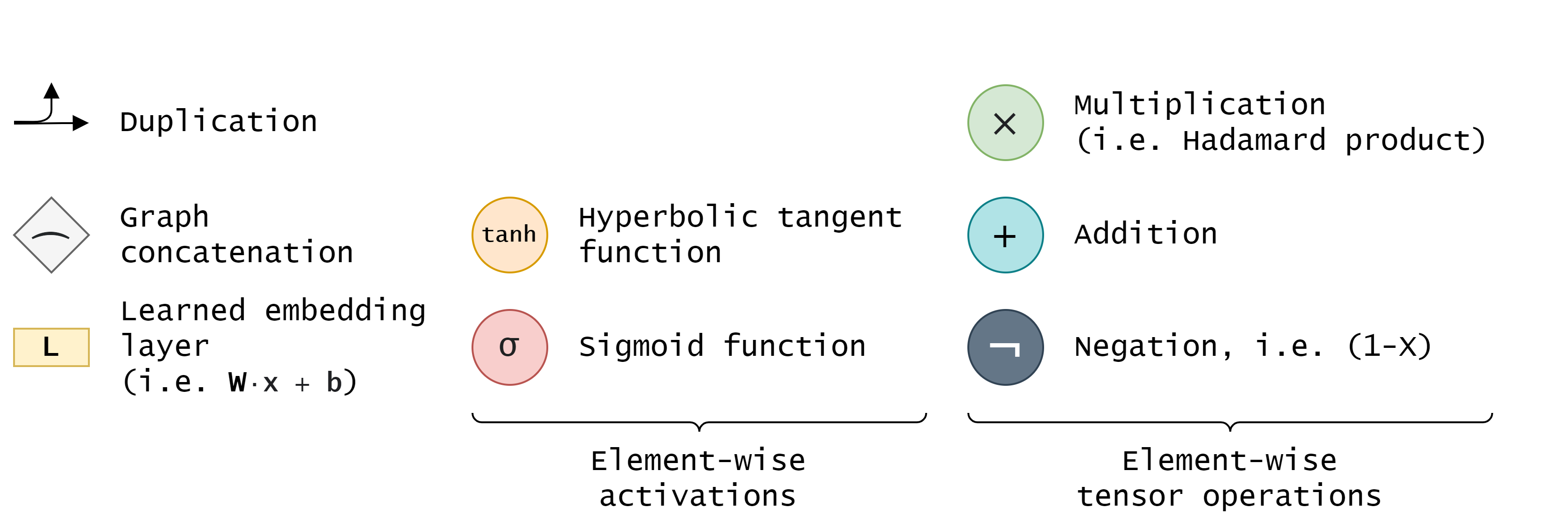}
    }
    \\

    \caption{The processing approaches for graph-based sequential data, including the overarching approach (a), simple RNN cells (b), GLSTMs (c), and graph GRUs (d).}

    \label{fig:rnnextensions}
\end{figure}

As is the case with traditional sequential data, when processing each state of the sequence we want to consider not only the current state but also information from the previous states, as outlined in Figure~\ref{fig:rnnextensions} (a). A simple solution to this challenge might be to simply concatenate the graph emebddings of previous states to the features of the current state (as in Figure~\ref{fig:rnnextensions} (b)), but such approaches do not capture long term dependencies in the data. In this section, we outline how existing solutions from traditional DL --- such as Long Short-Term Memory Networks (LSTMs) and Gated Recurrent Units (GRUs) (outlined in Figure~\ref{fig:rnnextensions}) --- can be extended to the graph domain.

\label{s:glstm}
\textbf{Graph LSTMs} (GLSTMs) make use of LSTM cells that have been adapted to operate on graph based data. Whereas the aforementioned recurrent modules (Figure~\ref{fig:rnnextensions} (b)) employ a simple concatenation strategy, GLSTMs ensure that long-term dependencies can be encoded in the LSTM's `cell state' (Figure~\ref{fig:rnnextensions} (c)). This alleviates the vanishing gradient problem where long-term dependency signals are exponentially reduced when backpropagated throughout the network \cite{hochreiter1991untersuchungen, hochreiter2001gradient}. 

GLSTM cells achieves this through \textbf{four} key processing elements which learn to calculate useful quantities based on the previous state's embedding and the input from the current state (as illustrated in Figure~\ref{fig:rnnextensions} (c)). 

\begin{enumerate}
    \item The \textbf{forget gate}, which uses $L_f$ to extract values in the range [0,1], representing if elements in the previous cell's state should be `forgotten' (0) or retained (1). 
    \item The \textbf{input gate}, which uses $L_i$ to extract values in the range [0,1], indicating the amount of the modulated input which will be added to this cell's cell state. 
    \item The \textbf{input modulation gate}, which uses $L_n$ to extract values in the range [-1,1], representing learned information from this cell's input.
    \item The \textbf{output gate}, which uses $L_o$ to calculate values in the range [0,1], indicating which parts of the cell state should be output as this cell's hidden state.
\end{enumerate}

To use GLSTMs, we need to define all the operators in Figure~\ref{fig:rnnextensions} (e). Since a graph \Glong can be thought of as a variably sized set of vertices and edges, we can define graph concatenation as the separate concatenation of vertex features and edge features, where some null padding is used to ensure that the resultant tensor is of a fixed-size. This can be achieved by defining some `max number' of vertices for the input graphs. If the input signal for the GLSTM cell has a fixed size, then all other operators can be interpreted as traditional tensor operations, and the entire process is differentiable when it comes to backpropagation.

\begin{mdframed}[style=examplebox]
\vspace{-0.3cm}
\subsection*{The Role of Recurrent Transitions in RGNNs for Graph Classification}
\vspace{-1mm}
\label{ss:ex1}
\small

\edit{In this independent example, we investigate social networks, which represent a rich source of graph data. Due} to the popularity of social networking applications, accurate user and community classifications have become exceedingly important for the purpose of analysis, marketing, and influencing. In this example, we look at how the recurrent application of a transition function aids in making predictions on the graph domain, namely, in graph classification. 

\begin{wrapfigure}{r}{3.3cm}
    \vspace{-0.45cm}
    \fbox{\includegraphics[width=3.05cm]{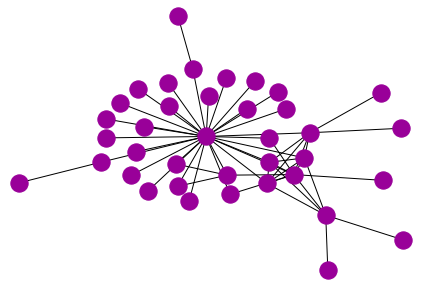}}
    \vspace{-5mm}
    \caption{A web developer group. $\numV = 38$ and $\numE = 110$.}
    \label{fig:ghstarweb}
    \vspace{1mm}
    \fbox{\includegraphics[width=3.05cm]{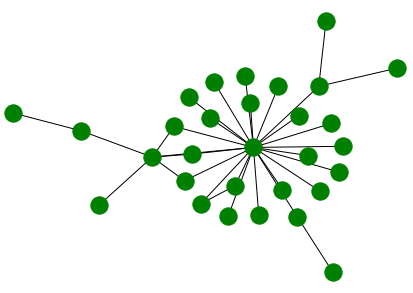}}
    \vspace{-5mm}
    \caption{A machine learning developer group. $\numV = 30$ and $\numE = 66$.}
    \label{fig:ghstarml}
\end{wrapfigure}

\vspace{-0.25cm}
\subsection*{Dataset}
\vspace{-0.11cm}

We will be using the GitHub Stargazer's dataset \cite{rozemberczki2020api} \href{http://snap.stanford.edu/data/github_stargazers.html}{(\underline{available here})}. GitHub is a code sharing platform with social network elements. \textit{Each} of the $12725$ graphs is defined by a group of users (vertices) and their mutual following relationships (undirected edges). Each graph is classified as either a web development group, or a machine learning development group. There are \textbf{no vertex or edge features} --- all predictions are made entirely from the structure of the graph.

\vspace{-0.25cm}
\subsection*{Algorithms}
\vspace{-0.11cm}

Rather than use a true RGNN which applies a transition function to hidden states until some convergence criteria is reached, we will instead experiment with limited applications of the transition function. The transition function is a simple message passing aggregator which applies a learned set of weights to create size $16$ hidden vector representations. We will see how the prediction task is affected by applying this transition function $1, 2, 4,$ and $8$ times before feeding the hidden representations to an output function for graph classification. We train on $8096$ graphs for $16$ epochs and test on $2048$ graphs for each architecture. 

\vspace{-0.25cm}
\subsection*{Results and Discussion}
\vspace{-0.11cm}

As expected, successive transition functions result in more discriminative features being calculated, thus resulting in a more discriminative final representation of the graph (analagous to more convolutional layers in a CNN).

\begingroup
\setlength{\intextsep}{0pt}%
\setlength{\columnsep}{0pt}%
\begin{wraptable}{l}{5cm}
    \vspace{-0.4cm}
    \begin{tabular}{|l|c|c|}
        \hline
        Algorithm & Acc. (\%) & AUC\\
        \hline
        x1 transition & 52\% & 0.5109 \\
        x2 transition & 55\% & 0.5440 \\
        x4 transition & 56\% & 0.5547 \\
        x8 transition & \textbf{64}\% & \textbf{0.6377} \\
        \hline
    \end{tabular}
    \caption{The effect of repeated transition function applications on graph classification performance}
    \label{tab:ghstarresults}
\end{wraptable}
\endgroup

\noindent In fact, we can see that the final hidden representations become more linearly separable (see TSNE visualizations in Figure~\ref{fig:tsne}), thus, when they are fed to the output function --- a linear classifier --- the predicted classifications are more often correct. This is a difficult task since there are no vertex or edge features. State of the art approaches achieved the following mean AUC values averaged over 100 random train/test splits for the same dataset and task: GL2Vec \cite{chen2019GL2vecGE} --- 0.551, Graph2Vec \cite{narayanan2017graph2vec} --- 0.585, SF \cite{lara2018baseline} --- 0.558, \textbf{FGSD \cite{verma2017hunt} --- 0.656}.

\vspace{-3mm}
\begin{figure}[H]
    \fbox{\begin{subfigure}{.24\textwidth}
        \centering
        \includegraphics[width=.95\textwidth]{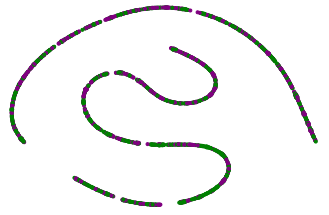}
    \end{subfigure}%
    \hspace{0cm}
    \begin{subfigure}{.24\textwidth}
      \centering
      \includegraphics[width=.95\textwidth]{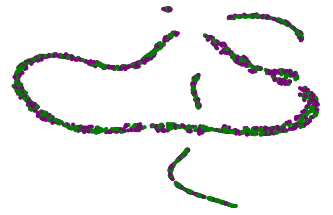}
    \end{subfigure}
    \hspace{0cm}
    \begin{subfigure}{.24\textwidth}
      \centering
      \includegraphics[width=.95\textwidth]{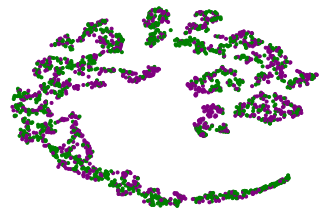}
    \end{subfigure}%
    \hspace{0cm}
    \begin{subfigure}{.24\textwidth}
      \centering
      \includegraphics[width=.95\textwidth]{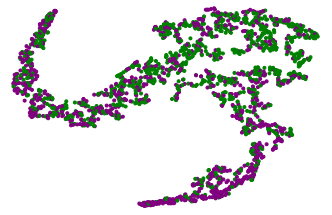}
    \end{subfigure}%
    }
    \vspace{-4mm}\caption{TSNE renderings of final hidden graph representations for the x1, x2, x4, x8 hidden layer networks. Note that with more applications of the transition function (equivalent to more layers in a NN) the final hidden representations of the input graphs become \textit{more} linearly separable into their classes (hence why they are able to be better classified using only a linear classifier).} 
    \label{fig:tsne}
\end{figure}

\vspace{-5mm}\noindent Here, our transition function \fntrans was a `feedforward NN' with just one layer, so more advanced NNs (or other) implementations of \fntrans might result in more performant RGNNs. As more rounds of transition function were applied to our hidden states, the performance --- and required computation --- increased. Ensuring a consistent number of transition function applications is key in developing simplified GNN architectures, and in reducing the amount of computation required in the transition stage. We will explore how this improved concept is realised through CGNNs in Section~\ref{s:cgnnspatial}.

\end{mdframed}




\label{s:gated}

\textbf{Gated Recurrent Units} (GRUs) provide a less computationally expensive alternative to GLSTMs by removing the need to calculate a cell state in each cell. As such, GRUs have \textbf{three} learnable weight matrices (as illustrated in Figure~\ref{fig:rnnextensions} (d)) which serve similar functions to the four learnable weight matrices in GLSTMs. Again, GRUs require some definition of \textit{graph concatenation}. 


\begin{enumerate}
    \item The \textbf{reset gate} $L_r$ determines how much information from to `forget' or `retain' when calculating the new information to add to the hidden state from the current state. 
    \item The \textbf{update gate} $L_u$ determines what information to `forget' or `retain' from the previous hidden state.
    \item The \textbf{candidate gate} $L_n$ determines what information from the reset input will contribute to the next hidden state.
\end{enumerate}

GRUs are well suited to sequential data when repeating patterns are less frequent, whereas LSTM cells perform well in cases where more frequent pattern information needs to be captured. LSTMs also have a tendency to overfit when compared to GRUs, and as such GRUs outperform LSTM cells when the sample size is low \cite{gruber2020gru}.






\subsection{\edit{Advantages, Disadvantages, and Applications}}


In this section, we have explained the forward pass that allows a RGNN to produce useful predictions over graph input data. During the forward pass, a transition function \fntrans is recursively applied to an input graph to create high level features for each neighborhood. The repeated application of \fntrans ensures that at iteration $k$, an embedding \hidx{i}{k} includes information from vertices $k$ edges away from $\vidx{i}$. These high-level features can be fed to an output function to solve downstream tasks. During the backward pass, the parameters for the NNs \fntrans and \fnoutput are updated with respect to a loss which is backpropagated through the computation graph defined in the forward pass. Recurrent processing units can also refer to approaches for handling graph-based sequential data, which include graph-based extensions to LSTMs and GRUs.  

\edit{In actuality, the formulation for calculating embeddings provided in Equation~\ref{eq:rgnn} represents only one approach to calculating embeddings. This approach will be contextualised in Section~\ref{s:cgnn}, where a broader perspective on calculating useful embeddings will be introduced.}

\edit{While RGNNs offer a simple approach to working with generic graphs, they have a number of shortcomings. Namely, the shared transition function \fntrans means that the same weights are being used to extract features in successive iterations, which many not be ideal for deep learning scenarios where the relationships between low level features (earlier in the network) are different to the relationships between high level features (later in the network). Moreover, since RGNNs iterate until convergence, they have variable length encoding networks, which can add implementation complexities. In the next section, we will discuss how these issues can be alleviated by developing formal definitions of convolution in the graph domain.  } 

\section{Convolutional Graph Neural Networks}
\label{s:cgnn}

\edit{Convolutional NNs have achieved state-of-the-art performance on predictive tasks involving images. By convolving a learned kernel of weights with an input image, CNNs extract features of interest based on their visual appearance --- regardless of their locality in the image. Since images are just a special case of graphs (see Figure~\ref{fig:abstractgeneral} (a)), a generalised convolution operator can be defined for the graph domain, thus bringing the following desirable properties to GNNs: }

\begingroup
\setlength\arraycolsep{2pt}
\renewcommand*{\arraystretch}{1.1}
\begin{table}[H]
    \begin{tabular}{p{4.0cm} p{9.0cm} }
        \toprule
        Approach & Applications \\
        \midrule
        
        RNNs (early work) \cite{micheli2001analysis} & Quantitative structure-activity relationship analysis. \\
        
        RNNs (early work) \cite{bianchini2002recursive} & Various, including localisation of objects in images. \\
        
        RGNNs (early work) \cite{scarselli2009gnn} & Various, including subgraph matching, the mutagenesis problem, and web page ranking. \\
        
        RGNNs (Neural Networks for Graphs) \cite{micheli2009neural} & Quantitative structure-activity relationship analysis of alkanes, and classification of general cyclic / acyclic graphs \\
        
        RGNNs \& RNNs (a comparison) \cite{di2006comparison} & $4$-class image classification  \\

        \midrule 
        
        Geometric Deep Learning algorithms (incl. RGNNs) \cite{bronstein2016geometric} & Graphs, grids, groups, geodesics, gauges, point clouds, meshes, and manifolds. Specific investigations include computer graphics, chemistry (e.g. drug design), protein biology, recommender systems, social networks, traffic forcasting, etc. \\
        
        RGNN pretraining \cite{hu2020pretrain} & Molecular property prediction, protein function prediction, binary graph classification, etc. \\
        
        RGNNs benchmarking \cite{loukas2019depthwidth} & Cycle detection, and exploring what RGNNs can and cannot learn. \\ 
        
        Natural Graph Networks (NGNs) \cite{haan2020natural} & Graph classification (bioinformatics and social networks). \\
        
        \midrule 
        
        GLSTMs \cite{zeng2021deep} & Airport delay prediction (with \numV$=325$). \\ 
        
        GLSTMs (using differential entropy) \cite{yin2021eeg} & Emotion classification from electroencephalogram (EEG) analysis (graphs calculated from K-nearest neighbor algorithms). \\ 
        GLSTMs \cite{lu2020lstm} & Speed prediction of road traffic throughout a directed road network (vertices are road segments, and edges are direct links between them). \\
        
        GLSTMs (with spatiotemporal graph convolution) \cite{pan2021driver} & Real-time distracted driver behaviour classification (i.e., based on the human pose graph \cite{fang2017alphapose} from a sequence of video frames, is the driver drinking, texting, performing normal driving, etc.). Other techniques for this problem include \cite{li2020graph, si2019attention}. \\
        
        LSTM-Q (i.e. fusion of RL with a bidirectional LSTM) for graphs \cite{chen2021graph} &  Connected autonomous vehicle network analysis for controlling agent movement (in a multi-lane road corridor).  \\
        
        
        \midrule 
        
        Graph GRUs \cite{li2015gated} & Computer program verification. \\ 
        
        Graph GRUs \cite{harl2020explainable} & Explainable predictive business process monitoring. \\
        
        Graph GRUs \cite{beck2018graph} & NLP as a graph to sequence problem (leveraging structural information in language). \\
        
        Graph GRUs \cite{ruiz2019gated, ruiz2020gated} & Gating for vertices and edges. Key applications include earthquake epicentre placement and synthetic regression problems. \\ 
        
        Symmetric Graph GRUs \cite{lukovnikov2020improving} & Improved long term dependency performance on synthetic tasks. \\

        \bottomrule
    \end{tabular}
\caption{A selection of works which use recurrent GNN techniques such as those discussed in this section. }
\label{tab:rgnnapps}
\end{table}
\endgroup







\edit{\begin{enumerate}
    \item \textbf{Locality}: learned feature extraction weights should be localised. They should only consider the information within a given neighborhood, and they should be applicable throughout the input graph.
    \item \textbf{Scalability}: the learning of these feature extractors should be scalable, i.e., the number \edittwo{of} learnable parameters should be independent of \numV. Preferably the operator should be `stackable', so that models can be built from successive independent layers, rather than requiring repeated iteration until convergence as with RGNNs in Section~\ref{s:rgnn}. Computation complexities should be bounded where possible.
    \item \textbf{Interpretability}: the convolutional operation should (preferably) be grounded in some mathematical or physical interpretation, and its mechanics should be intuitive to understand.  
\end{enumerate}}

\subsection{What is Convolution?}

\edit{We define convolution generally as an operation whereby \textbf{an output is derived from \textit{two} given inputs by integration or summation, which expresses how the one is modified by the other}. }

\begin{figure}[H]
    \centering
    \subfloat[A convolutional operation of 2D matrices. This process is used throughout computer vision and in CNNs. The convolutional operation here has a stride of $2$ pixels. The given filter is applied in the red, green, blue, and then purple positions. At each position each element of the filter is multiplied with the corresponding element in the input (i.e., the Hadamard product) and the results are summed, producing a single element in the output. For clarity, this multiplication and summing process is illustrated for the purple position. In the case of this image the filter is a standard sharpening filter used in image analysis.]{
        \includegraphics[width=0.58\textwidth]{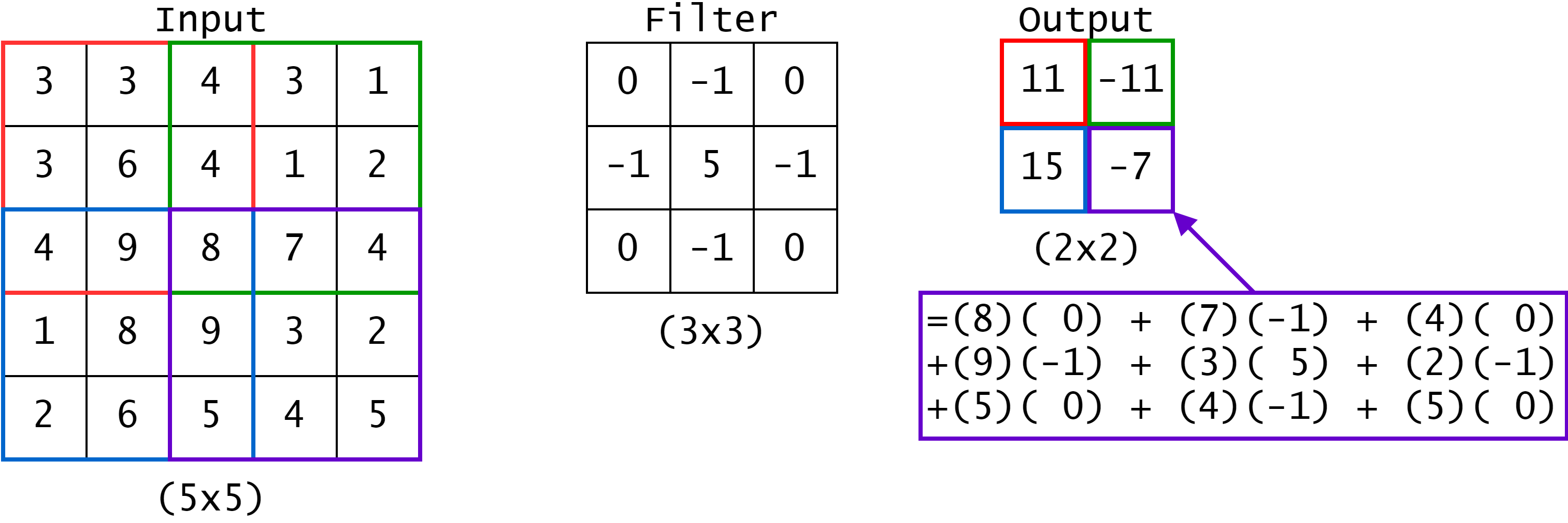}
    }\qquad
    \subfloat[Three neighborhoods in a given graph (designated by dotted boxes), with each one defined by a central vertex (designated by a correspondingly coloured circle). Each neighborhood in the graph is aggregated into a feature vector (i.e., and embedding) centered at each vertex, thus allowing the process to repeat for multiple layers. 
    ]{
        \includegraphics[width=0.30\textwidth]{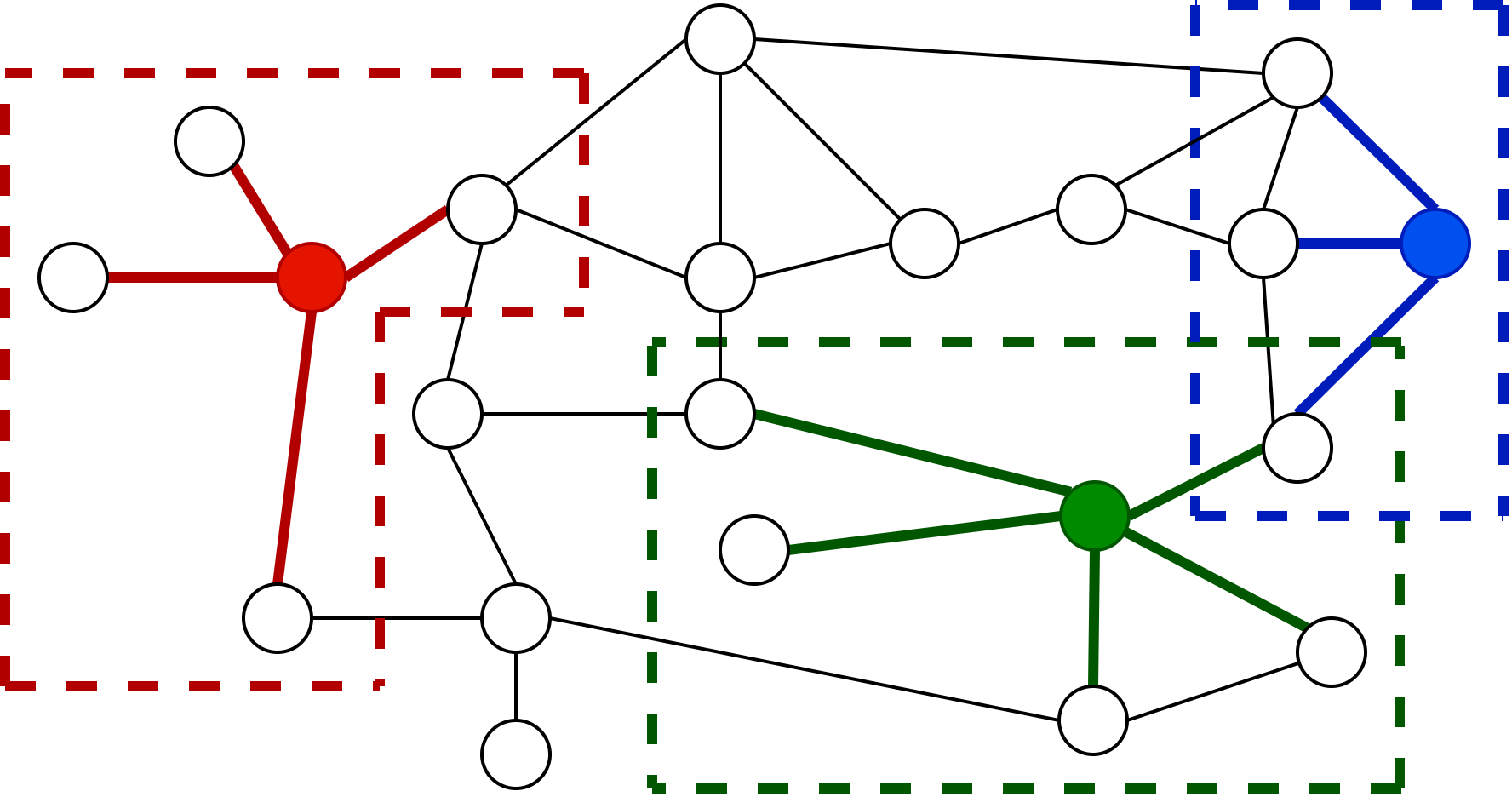}
    }
    \caption{\todo{A comparison of image-based and graph-based spatial convolution techniques. Both techniques create embeddings centered around pixels / vertices, and the output of both techniques describes how the input is modified by the filter. Images best viewed in colour.}}
    \label{fig:tradconv}
\end{figure}

Convolution in CNNs involves two matrix inputs, one is the previous layer of activations, and the other is a matrix $W \times H$ of learned weights, which is `slid' across the activation matrix, aggregating each $W \times H$ region using a simple linear combination (see Figure~\ref{fig:tradconv} (a)). In the spatial graph domain, it seems that this type of convolution is not well defined \cite{shuman2013emerging}; the convolution of a rigid matrix of learned weights must occur on a rigid structure of activation. How do we reconcile convolutions on unstructured inputs such as graphs? 


Note that at no point during our \edit{general} definition of convolution was the structure of the given inputs alluded to. In fact, convolutional operations can be applied to continuous functions (e.g., audio recordings and other signals), N-dimensional discrete tensors (e.g., semantic vectors in 1D, and images in 2D), and so on. During convolution, one input is typically interpreted as a filter (or \textit{kernel}) being applied to the other input, and we will adopt this language throughout this section. Specific filters can be utilised to perform specific tasks: in the case of audio recordings, high pass filters can be used to filter out low frequency signals, and in the case of images, certain filters can be used to increase contrast, sharpen, or blur images. In our previous example of CNNs, filters are learned rather than designed.  

\subsection{\edit{Spatial Approaches}}
\label{s:cgnnspatial}

\edit{One might consider the early RGNNs described in Section~\ref{s:rgnn} as using convolutional operations. In fact, these methods meet the criteria of locality, scalability, and interpretability. Firstly, Equation~\ref{eq:rgnn} only operates over the neighborhood of the central vertex \vidx{i}, and can be applied on any neighborhood in the graph due to its invariance to permutation and neighborhood size. Secondly, the NN \fntrans is dependent on a fixed number of weights, and has a fixed input and output that is independent of \numV. Finally, the convolution operation is immediately interpretable as a generalisation of image based convolution: in image based convolution neighboring pixel values are aggregated to produce embeddings, in graph-based spatial convolution neighboring vertex features are aggregated to produce embeddings (see Figure~\ref{fig:tradconv}). This type of graph convolution is referred to as the \textbf{spatial} graph convolutional operation, since spatial connectivity is used to retrieve the neighborhoods in this process.}




\edit{Although the RGNN technique meets the definition of spatial convolution, there are numerous improvements in the literature.} For example, the choice of aggregation function is not trivial --- different aggregation functions can have notable effects on performance and computational cost.







A notable framework that investigated aggregator selection is the \textbf{GraphSAGE} framework \cite{hamilton2017graphsage}, which demonstrated that learned aggregators can outperform simpler aggregation functions (such as taking the mean of embeddings) and thus can create more discriminative, powerful vertex embeddings. Regardless of the aggregation function, GraphSAGE works by computing embeddings based on the central vertex and an aggregation of its neighborhood (see Equation~\ref{eq:graphsage}). By including the central vertex, it ensures that vertices with near identical neighborhoods have different embeddings. GraphSAGE has since been outperformed on accepted benchmarks \cite{dwivedi2020benchmarking} by other frameworks \cite{bresson2017residual}, but the framework is still competitive and can be used to explore the concept of learned aggregators (see Section~\ref{ss:ex2}). 

\begin{equation}
    \hidx{i}{k} = \sigma(\textbf{W}\textnormal{concat}(\hidx{i}{k-1}, \textnormal{aggregate}(\hidx{j}{k-1}\; \forall j \in\textnormal{\nbh{\vidx{i}}})))
    \label{eq:graphsage}
\end{equation}

Alternatively, \textbf{Message Passing Neural Networks} (MPNNs) compute \textit{directional} messages between vertices with a message function that is dependent on the source vertex, the destination vertex, and the edge connecting them \cite{gilmer2017neural}. Rather than aggregate the neighbor's features and concatenating them with the central vertex's features as in GraphSAGE, MPNNs sum the incoming messages, and pass the result to a readout function alongside the central vertex's features (see Equation~\ref{eq:mpnn}). Both the message function and readout function can be implemented with simple NNs in practice. This generalises the concepts outlined in Equation~\ref{eq:rgnn}, and allows for more meaningful patterns to be identified by the learned functions.
\label{s:mpnn}

\begin{equation}
    \hidx{i}{k} = f_{readout}(\hidx{i}{k-1}, \sum_{j \in\textnormal{\nbh{\vidx{i}}}} m_{ij}^{k})), \, \textnormal{where $m_{ij}^{k}=f_{message}($\hidx{i}{k}, \hidx{j}{k}, \eidx{i}{j}$)$}
    \label{eq:mpnn}
\end{equation}

One of the most popular spatial convolution methods is \textbf{Graph Convolutional Networks} (GCNs), which produce embeddings by summing features extracted from each neighboring vertex and then applying non-linearity \cite{wu2019simpgcn}. These methods are highly scalable, local, and furthermore, they can be `stacked' to produce layers in a CGNN. Each of these features is normalised based on the relative neighborhood scales of the current and neighbor vertex, thus ensuring that embeddings do not 'explode' in scale during the forward pass.

\begin{equation}
    \hidx{i}{k} = \sigma(\sum_{j \in\textnormal{\nbh{\vidx{i}}}}   \frac{\textbf{W} \hidx{j}{k-1}}{\sqrt{ \magline{\nbh{\vidx{i}}} \magline{\nbh{\vidx{j}}} }})
    \label{eq:gcn}
\end{equation}

\textbf{Graph Attention Networks} (GATs) extend GCNs: instead of using the size of the neighborhoods to weight the importance of \vidx{i} to \vidx{j}, they implicitly calculate this weighting based on the normalised product of an attention mechanism \cite{vaswani2017attention}. In this case, the attention mechanism is dependent on the embeddings of two vertices and the edge between them. Vertices are constrained to only be able to attend to neighboring vertices, thus localising the filters. GATs are stabilised during training using multi-head attention and regularisation, and are considered less general than MPNNs \cite{velivckovic2017graph}. Although GATs limit the attention mechanism to the direct neighborhood, the scalability to large graphs is not guaranteed, as attention mechanisms have compute complexities that grow quadratically with the number of vertices being considered.
\label{s:gat}

\begin{equation}
    \hidx{i}{k} = \sigma(\sum_{j \in\textnormal{\nbh{\vidx{i}}}} \alpha_{ij} \textbf{W} \hidx{j}{k-1}), \;\textnormal{where } \alpha_{ij}=\frac{e^{\textnormal{att}(\hidx{i}{k-1}, \hidx{j}{k-1}, \eidx{i}{j})}}{\sum_{l\in\textnormal{\nbh{\vidx{i}}}} e^{\textnormal{att}(\hidx{i}{k-1}, \hidx{l}{k-1}, \eidx{i}{l})}}
\end{equation}

Interestingly, all of these approaches consider information from the direct neighborhood and the previous embeddings, aggregate this information in some symmetric fashion, apply learned weights to calculate more complex features, and `activate' these results in some way to produce an embedding that captures non-linear relationships.

\subsection{\edit{Spectral Approaches}}
\label{s:cgnnspectral}

\edit{In this section, we discuss another class of convolution approaches that evolved from the perspective of Graph Signal Processing (GSP) \cite{stankovic2019understandinggsp, shuman2013emerging}. These methods are attractive as they are well grounded in a formal definition of convolution, and can be directly interpreted as signal processing techniques in the domain of graph structured data.}

The path to defining spectral graph convolution is described by the following series of statements.

\begin{equation}
    \label{eq:conv}
    (f * g)(t) = \int_{-\infty}^{+\infty} f(u)g(t - u) du
\end{equation}

\begin{enumerate}
    \item Defining a convolutional operator in the graph domain is desirable (as motivated in Section~\ref{s:cgnnspatial}).
    \item From a signal processing perspective, the convolution operator is defined as in Equation~\ref{eq:conv}. In other words, it is the integral of the product of a reversed and translated filter ($g(t-u)$) and an input function ($f(u)$). To define this in the graph domain, a translation operator needs to be defined for graphs.
    \item By Parseval's theorem, multiplication in the frequency domain (frequency space) corresponds to translation in the spatial domain (vertex space) \cite{li2019convolutiontheorem}. Formally defining spatial translation in the graph domain requires a method to convert graphs between the vertex and frequency space. 
    \item The eigenfunctions of the Laplacian define a basis in frequency space, so a formal definition of the graph Laplacian is required to develop spectral graph convolutions.
\end{enumerate}













\newpage
\begin{mdframed}[style=examplebox]
\vspace{-0.3cm}
\subsection*{Using GraphSAGE to Generate Embeddings for Unseen Data}
\label{ss:ex2}
\small

\vspace{-0.1cm}

The GraphSAGE (\textbf{SA}mple and aggre\textbf{G}a\textbf{t}E) algorithm \cite{hamilton2017graphsage} emerged in 2017 as a method for not only learning useful vertex embeddings, but also for predicting vertex embeddings on \textit{unseen} vertices. This allows powerful high-level feature vectors to be produced for vertices which were not seen at train time; enabling us to effectively work with dynamic graphs, or very large graphs (>$100,000$ vertices).

\vspace{-0.3cm}
\subsection*{Dataset}
\vspace{-0.1cm}

In this example we use the Cora dataset (see Figure~\ref{fig:cora_graphsage}) as provided by the deep learning library \textit{DGL} \cite{wang2019dgl}. The Cora dataset is oft considered `the MNIST of graph-based learning' and consists of $2708$ scientific publications (vertices), each classified into one of seven \change{subfields in AI} (or classes). Each vertex has a $1433$ element binary feature vector, which indicates if each of the $1433$ designated words appeared in the

\begin{wrapfigure}{r}{5.2cm}
    \vspace{-3mm}
    \fbox{\includegraphics[width=4.9cm]{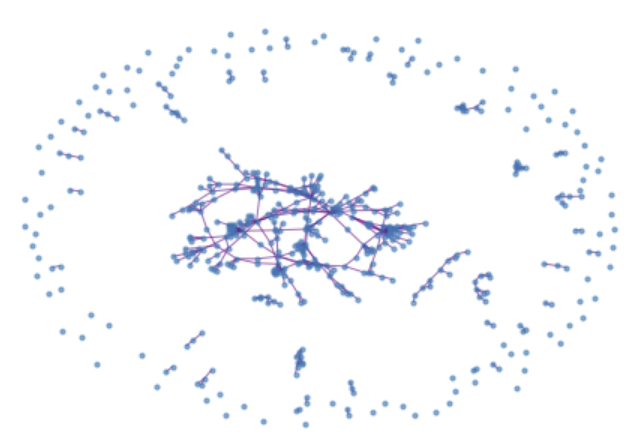}}
    \vspace{-3mm}
    \caption{A subgraph of the Cora dataset. The full Cora graph has $\numV = 2708$ and $\numE = 5429$. Note the many vertices with few incident edges (low degree) as compared to the few vertices with many incident edges (high degree). }
    \label{fig:cora_graphsage}
\end{wrapfigure}

\noindent publication.



\vspace{-0.3cm}
\noindent\subsection*{What is GraphSAGE?}
\vspace{-0.1cm}


\noindent GraphSAGE operates on a simple assumption: \textit{vertices with similar neighborhoods should have similar embeddings}. In this way, when calculating a vertex's embedding, GraphSAGE considers the vertex's neighbors' embeddings. The function which produces the embedding from the neighbors' embeddings is learned, rather than the embedding being learned directly. Consequently, this method is \textbf{not} \textit{transductive}, it is \textit{inductive}, in that it generates general rules which can be applied to unseen vertices, rather than reasoning from specific training cases to specific test cases.

Importantly, the GraphSAGE loss function is unsupervised, and uses two distinct terms to ensure that neighboring vertices have similar embeddings and distant or disconnected vertices have embedding vectors which are numerically far apart. This ensures that the calculated vertex embeddings are highly discriminative. 




\vspace{-0.3cm}
\subsection*{Architectures}
\vspace{-0.1cm}

In this worked example, we experiment by changing the aggregator functions used in each GNN and observe how this affects our overall test accuracy. In all experiments, we use $2$ hidden GraphSAGE convolution layers, $16$ hidden channels (i.e., embedding vectors have $16$ elements), and we train for $120$ epochs before testing our vertex classification accuracy. We consider the \textbf{mean}, \textbf{pool}, and \textbf{LSTM} (long short-term memory) aggregator functions.

The \textbf{mean} aggregator function sums the neighborhood's vertex embeddings and then divides the result by the number of vertices considered. The \textbf{pool} aggregator function is actually a single fully connected layer with a non-linear activation function which then has its output element-wise max pooled. The layer weights are learned, thus allowing the most import features to be selected. The \textbf{LSTM} aggregator function is an LSTM cell. Since LSTMs consider input sequence order, this means that different orders of neighbor embedding produce different vertex embeddings. To minimise this effect, the order of the input embeddings is randomised. This introduces the idea of aggregator \textbf{symmetry}; an aggregator function should produce a constant result, invariant to the order of the input embeddings.



    





\vspace{-0.3cm}
\subsection*{Results and Discussion}
\vspace{-0.1cm}
    
The mean, pool and LSTM aggregators score test accuracies of \bm{$66.0\%$}, \bm{$74.4\%$}, and \bm{$68.3\%$}, respectively. As expected, the learned pool and LSTM aggregators are \edit{more effective than the simple mean operation}, though they incur significant training overheads, and may not be suitable for smaller training graphs or graph datasets. Indeed, in the original GraphSAGE paper \cite{hamilton2017graphsage}, it was found that the LSTM and pool methods generally outperformed the mean and GCN aggregation methods across a range of datasets.




At the time of publication, GraphSAGE outperformed the state-of-the-art on a variety of graph-based tasks on common benchmark datasets. Since that time, a number of inductive learning variants of GraphSAGE have been developed, and their performance on benchmark datasets is regularly updated\footnote{ The state-of-the-art for vertex classification (Cora dataset): \url{https://paperswithcode.com/sota/node-classification-on-cora} }.

\end{mdframed}






\noindent \textbf{The Laplacian} is a second order differential operator that is calculated as the divergence of the gradient of a function in Euclidean space. The Laplacian occurs naturally in equations that model physical interactions, including but not limited to electromagnetism, heat diffusion, celestial mechanics, and pixel interactions in computer vision applications. Similarly, it arises naturally in the graph domain, where we are interested in the `diffusion of information' throughout a graph structure.

\begin{figure}[H]
    \centering
    
    \subfloat[An input function $f(x,y)$ (rendered as a purple surface plot) and its gradient $\nabla f(x,y) $ (rendered as an orange vector field above the surface plot). The gradient, at every point, denotes the direction which increases $f(x,y)$ the most. In other words, the orange arrows always point in the direction of a maxima in the purple surface plot. ]{
        \includegraphics[width=0.45\textwidth]{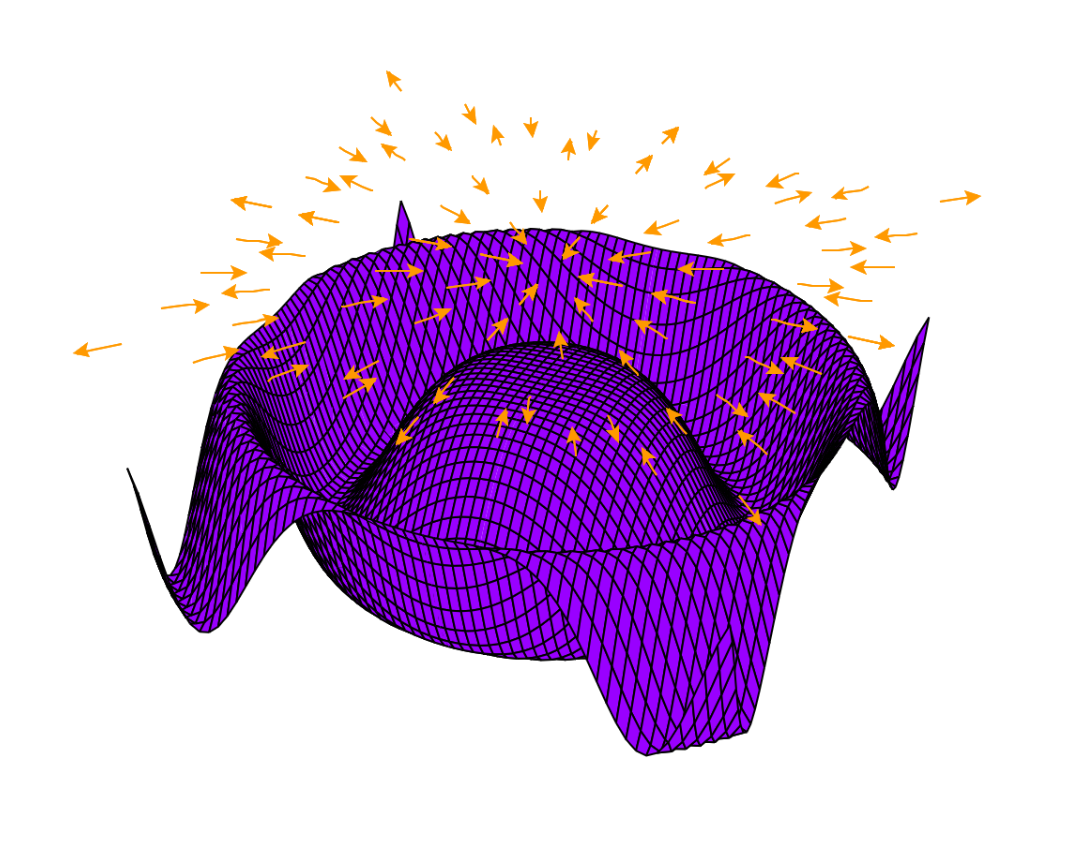}
    }\qquad
    \subfloat[The vector field from (a) $\nabla f(x,y) $ (rendered as an orange vector field above the surface plot) and its divergence $\nabla \cdot \nabla f(x,y)$ (rendered as a green surface plot). The divergence denotes how much every infinitesimal region of the vector field behaves like a source. In other words, it is a measure of the `outgoing flow' of the infinitesimal volume at each point. ]{
        \includegraphics[width=0.45\textwidth]{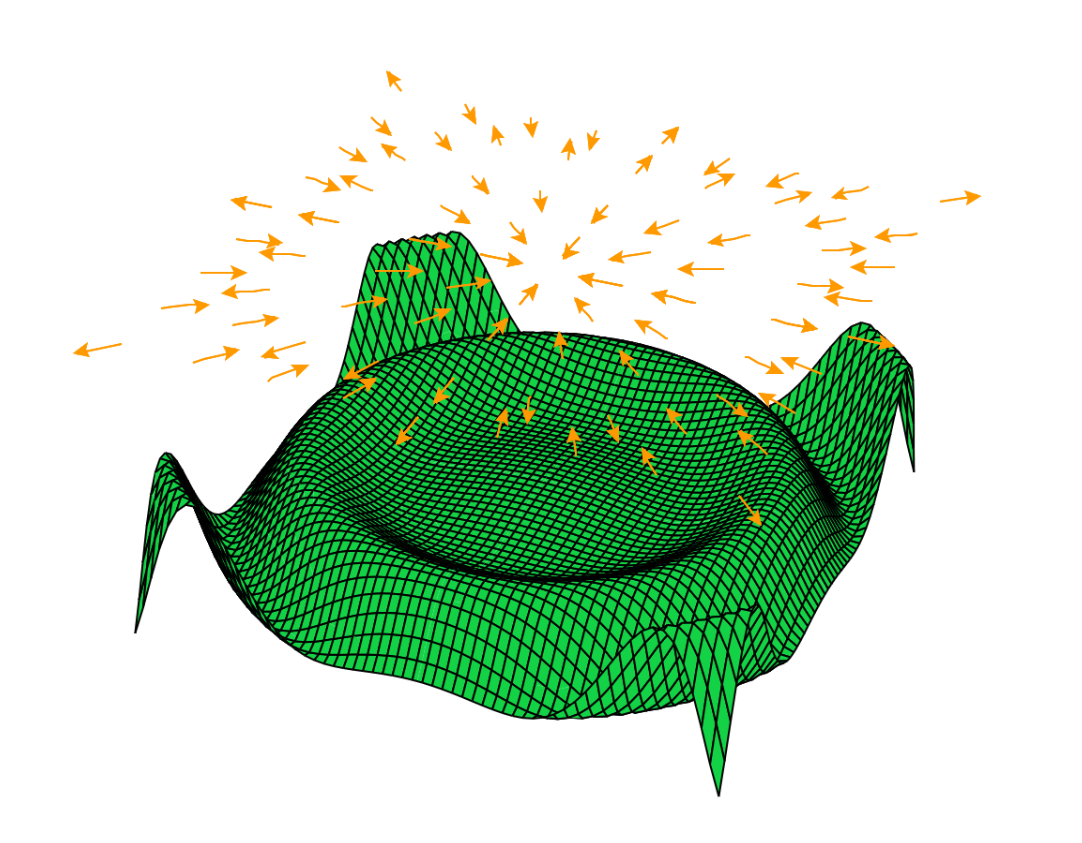}
    }
    \vspace{0cm}
    \\
    
    \caption{An input function $f(x,y):\real^2 \mapsto \real$ (a), its gradient $\nabla f(x,y):\real^2 \mapsto \real^2$ ((a) and (b)), and the divergence of its gradient $\nabla \cdot \nabla f(x,y):\real^2 \mapsto \real$ (b). The divergence of a function's gradient is known as the \textbf{Laplacian}, and it can be interpreted as measuring `how much' of a minimum each point is in the original function $f(x,y)$. The plots in (a) and (b) are an example of the entire calculation of the Laplacian; from scalar field to vector field (gradient), and then from vector field back to scalar field (divergence). The Laplacian is an analog of the second derivative, and is often denoted by $\nabla \cdot \nabla$, $\nabla^2$, or $\Delta$.}

    \label{fig:fngraddiv}
\end{figure}

More formally, if we define flux as the quantity passing outward through a surface, then the Laplacian represents the density of the flux \textit{of the gradient flow of a given function}. A step by step visualisation of the Laplacian's calculation is provided in Figure~\ref{fig:fngraddiv}. Note that the definition of the Laplacian is dependant on three things: \textbf{functions}, the \textbf{gradient} of a function, and the \textbf{divergence} of the gradient. Since we're seeking to define the Laplacian in the graph domain, we need to define how these constructs operate in the graph domain.

\begin{figure}[H]
    \centering
    
    \includegraphics[width=1.01\textwidth]{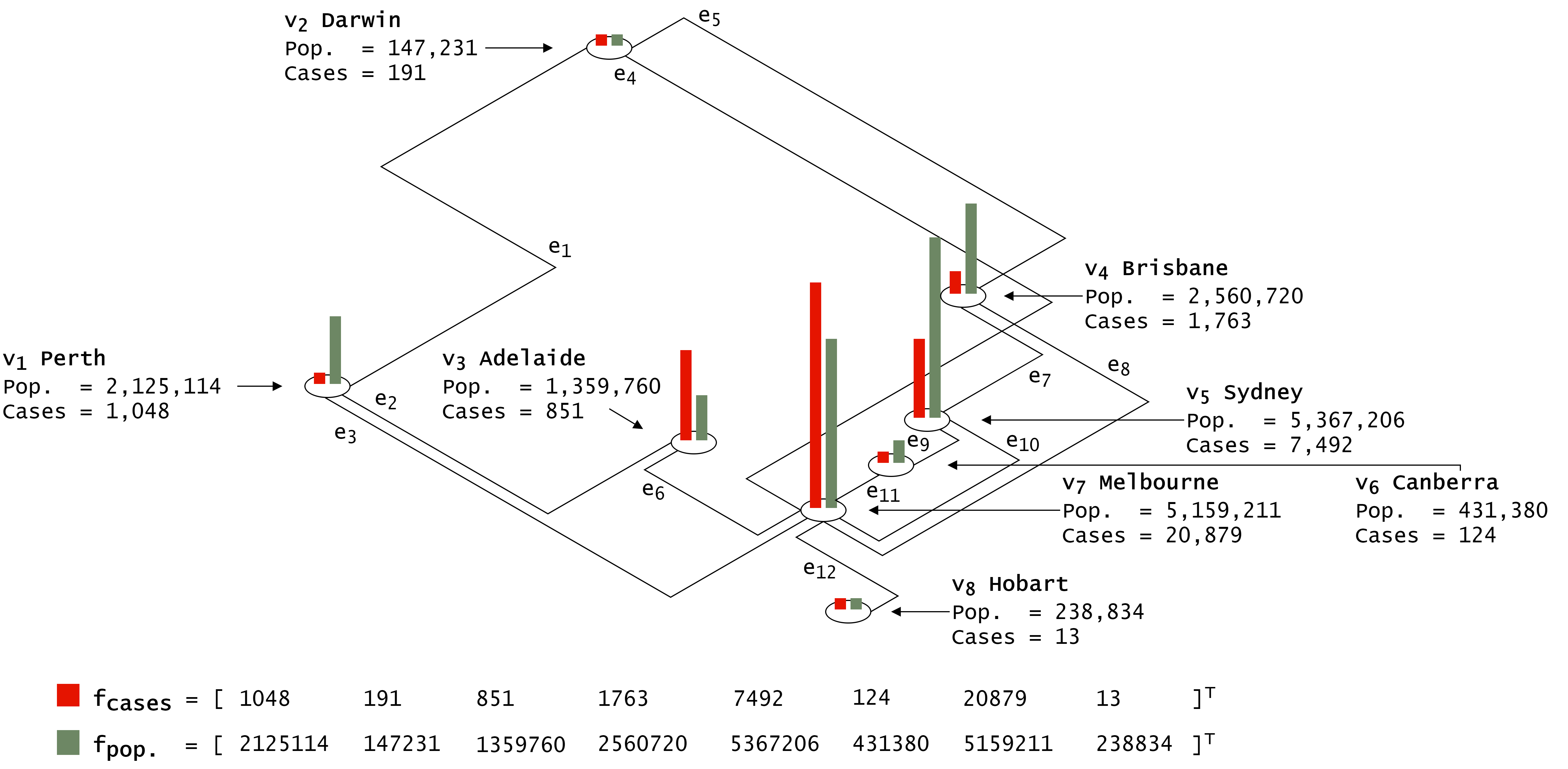}
    
    \caption{A graph representing Australia (\numV$=8$, \numE$=12$). Its vertices represent Australia's capital cities, and the edges between them represent common flight paths. Each vertex has two features, one representing the population, and another representing the total (statewide) cases of an infectious disease at those locations. Those two vertex feature vectors can be interpreted as the graph functions (also known as graph signals) $f_{\textnormal{cases}}$ and $f_{\textnormal{pop.}}$, which are rendered at the bottom of the figure. As an example, it may be of interest to investigate the propagation / diffusion of these graph signal quantities throughout the graph structure. }
    
    \label{fig:gfunction}
\end{figure}

\textbf{Functions in the graph domain} (referred to as graph signals in GSP) are a mapping from every vertex in a graph to a scalar value: $f(\Glong): \V \mapsto \real$. Multiple graph functions can be defined over a given graph, and we can interpret a single graph function as a single feature vector defined over the vertices in a graph. See Figure~\ref{fig:gfunction} for an example of a graph with two graph functions.

The \textbf{gradient of a function in the graph domain} describes the the direction and the rate of fastest increase of graph signals. In a graph structure, when we refer to `direction' we are referring to the edges of the graph; the avenues by which a graph function can change. For example, in Figure~\ref{fig:gfunction}, the graph functions are $8$-dimensional vectors (defined over the vertices), but the gradients of the functions for this graph are $12$-dimensional vectors (defined over the edges), and are calculated as in Equations~\ref{eq:gradient}. Refer to Table~\ref{tab:not} for a formal definition of the incident matrix \matincidence. 

\setcounter{MaxMatrixCols}{20}
\begin{equation}
\textnormal{
\noindent$\matincidence^\textnormal{T}=
    \begin{bmatrix*}[r]
        +1&-1&0&0&0&0&0&0\\
        +1&0&-1&0&0&0&0&0\\
        +1&0&0&0&0&0&-1&0\\
        0&+1&0&0&0&0&-1&0\\
        0&+1&0&-1&0&0&0&0\\
        0&0&+1&0&0&0&-1&0\\
        0&0&0&+1&-1&0&0&0\\
        0&0&0&+1&0&0&-1&0\\
        0&0&0&0&+1&-1&0&0\\
        0&0&0&0&+1&0&-1&0\\
        0&0&0&0&0&+1&-1&0\\
        0&0&0&0&0&0&+1&-1\\
    \end{bmatrix*}
$,\hspace{1mm}
$
f_{\textnormal{cases}}=\begin{bmatrix*}[r]
    1048 \\
    191 \\
    851 \\
    1763 \\
    7492 \\
    124 \\
    20879 \\
    13 \\
\end{bmatrix*}
$,\hspace{1mm}
$\nabla f_{\textnormal{cases}} = \matincidence^T f_{\textnormal{cases}} = \begin{bmatrix*}[r]
    +857\\ +\textbf{197} \\ -19831 \\ -20688 \\ -1572 \\ -20028 \\ -5729 \\ -19116 \\ +7368 \\ -13387 \\ -20755 \\ +\textbf{20866} \\
\end{bmatrix*}$
}
\label{eq:gradient}
\end{equation}

In Equations~\ref{eq:gradient}, the gradient vectors describe the difference in graph function value \textit{across} the vertices / \textit{along} the edges. Specifically, note that the largest magnitude value is 20866, and corresponds to \eidx{12}{}, the edge between Hobart and  Melbourne in Figure~\ref{fig:gfunction}. In other words, the greatest magnitude edge is between the city with the least cases and the city with the most cases. Similarly, the lowest magnitude edge is \eidx{2}{}; the edge between Perth and Adelaide, which has the least difference in cases.

The \textbf{divergence of a gradient function in the graph domain} describes the outward flux of the gradient function at every vertex. To continue with our example, we could interpret the divergence of the gradient function $f_{cases}$ as the outgoing `flow' of infectious disease cases from each capital city. Whereas the gradient function was defined over the graph's edges, the divergence of a gradient function is defined over the graph's vertices, and is calculated as in Equation~\ref{eq:div}. 

\begin{equation}
\textnormal{
$\nabla \cdot (\nabla f_{\textnormal{cases}}) = \matincidence   (\nabla f_{\textnormal{cases}}) = \matincidence  (\matincidence^T f_{\textnormal{cases}}) = \matincidence \matincidence^T f_{\textnormal{cases}} = \matlaplacian f_{\textnormal{cases}} = \begin{bmatrix*}[r]
    -18777\\ -23117\\ -20225\\ -23273\\ -290\\ -\textbf{28123}\\ +\textbf{134671}\\ -20866\\
\end{bmatrix*}$
}
\label{eq:div}
\end{equation}

The maximum value in the divergence vector for the infectious disease graph signal is $134671$, corresponding to Melbourne (the \addth{7} vertex). Again, this can be interpreted as the magnitude of the `source' of infectious disease cases from Melbourne. Contrastively, the minimum value is $-281123$, corresponding to Camberra, the largest `sink' of infections disease. 

Note as well that the dimensionality of the original graph function is $8$ --- corresponding to the vector space, its gradient's dimensionality is $12$ --- corresponding to its edge space, and the Laplacian's dimensionality is again $8$ --- corresponding to the vertex space. This mimics the calculation of the Laplacian in Figure~\ref{fig:fngraddiv}, where the original scalar field (representing the magnitude at each point) is converted to a vector field (representing direction), and then back to a scalar field (representing how each point acts as a source).


The \textbf{graph Laplacian} appears naturally in these calculations as a $\numV \times \numV$ matrix operator in the form $\matlaplacian=\matincidence\matincidence^T$ (see Equation~\ref{eq:div}). This corresponds to the formulation provided in Table~\ref{tab:not}, as shown in Equation~\ref{eq:lapex}, and this formulation is referred to as the combinatorial definition $\matlaplacian = \matdegree - \matweight$ (the normalised definition is defined as \matlaplacian$\!^{sn}$ \cite{defferrard2016convolutional}). The graph Laplacian is pervasive in the fields of GSP \cite{chung1997spectral}. 





\begin{equation}
\textnormal{
$\matlaplacian = \matincidence \matincidence^T = 
\begin{bmatrix*}[r]
 3 & -1 & -1 & 0 & 0 & 0 & -1 & 0 \\
-1 & 3 & 0 &-1  &0 & 0 &-1 & 0 \\
-1 & 0 & 2 & 0 & 0 & 0 &-1 & 0 \\
 0 &-1 & 0 & 3 &-1 & 0& -1 & 0 \\
 0 & 0 & 0 &-1 & 3 &-1 &-1 & 0 \\
 0 & 0 & 0 & 0 &-1 & 2 &-1 & 0 \\
-1 &-1 &-1 &-1 &-1 &-1 & 7 &-1 \\
 0 & 0 & 0 & 0 & 0 & 0 &-1 & 1 \\
\end{bmatrix*}
= \matdegree - \matweight
$
}
\label{eq:lapex}
\end{equation}



Since $\matlaplacian = \matdegree - \matweight$, the graph Laplacian must be a real ($\matlaplacian_{ij} \in \real, \hspace{1mm} \forall \hspace{1mm} 0 \leq i,j < \numV$) and symmetric ($\matlaplacian = \matlaplacian^T$) matrix. As such, it will have an \textbf{eigensystem} comprised of a set of \numV orthonormal eigenvectors, each associated with a single real eigenvalue \cite{shuman2013emerging}. We denote the \addth{i} eigenvector with $u_i$, and the associated eigenvalue with $\lambda_i$, each satisfying $\matlaplacian u_i = \lambda_i u_i$, where the eigenvectors $u_i$ are the \numV-dimensional columns in the matrix (Fourier basis) \mateigenvec. The Laplacian can be factored as three matrices such that $\matlaplacian = \mateigenvec \matlambda \mateigenvec^T$ through a process known as \textit{eigenvector decomposition}. A variety of algorithms exist for solving this kind of eigendecomposition problem (e.g. the QR algorithm and Singular Value Decomposition).

These eigenvectors form a basis in $\real^{\numV}$, and as such we can express any discrete graph function as a linear combination these eigenvectors. We define the \textit{graph Fourier transform} of any graph function / signal as $\hat{f} = \mateigenvec^T f \in \real^{\numV}$, and its inverse as $f = \mateigenvec \hat{f} \in \real^{\numV}$. To complete our goal of performing convolution in the spectral domain, we now complete the following steps.

\begin{enumerate}
    \item Convert the \addth{l} graph function into the frequency space (i.e., generate its graph Fourier transform). We do this through matrix multiplication with the transpose of the Fourier basis: $\mateigenvec^T f_l$. Note that multiplication with the eigenvector matrix is $O(N^2)$.
    
    \item Apply the corresponding \addth{l} learned filter in frequency space. If we define $\Theta_l$ as our \addth{l} learned filter (and a function of the eigenvalues of \matlaplacian), then this appears like so: $\Theta_l \mateigenvec^T f_l$.
    
    \item Convert the result back to vertex space by multiplying the result with the Fourier basis matrix. This completes the formulation defined in Equation~\ref{eq:eigenbasic}. By Parseval's theorem, multiplication applied in the frequency space corresponds to translation in vertex space, so the filter has been \textit{convolved} against the graph function \cite{li2019convolutiontheorem}.
    
\end{enumerate}

\begin{equation}
    \mateigenvec \Theta_l \mateigenvec^T f_l
    \label{eq:eigenbasic}
\end{equation}



This simple formulation has a number of downsides. Formostly, the approach is \textbf{not localised} --- it has \textit{global support} --- meaning that the filter is applied to all vertices (i.e., the entirety of the graph function). This means that useful filters aren't shared, and that the locality of graph structures is not being exploited. Secondly, it is \textbf{not scalable}; the number of learnable parameters grows with the size of the graph (not the scale of the filter) \cite{defferrard2016convolutional}, the $O(N^2)$ cost of matrix multiplication scales poorly to large graphs, and the $O(N^3)$ time complexity of QR-based eigendecomposition \cite{sleijpen2000jacobi} is prohibitive on large graphs. Moreover, directly computing this transform requires the diagonalisation of the Laplacian, and is \textit{infeasible} for large graphs (where \numV exceeds a few thousand vertices) \cite{hammond2011wavelets}. Finally, since the structure of the graph dictates the values of the Laplacian, \textbf{graphs with dynamic topologies can't use this method of convolution}.

\begin{equation}
    \Theta = \sum_{k=1}^{K} \theta_k T_k(\widetilde{\matlambda})
    \label{eq:eigenimp}
\end{equation}

To alleviate the \textbf{locality} issue \cite{estrach2014spectral} noted that the smoothing of filters in the frequency space would result in localisation in the vertex space. Instead of learning the filter directly, they formed the filter as a combination of smooth polynomial functions, and instead learned the coefficients to these polynomials. Since the Laplacian is a local operator affecting only direct neighbors of any given vertex, then a polynomial of degree $r$ affects vertices $r$-hops away. By approximating the spectral filter in this way (instead of directly learning it), spatial localisation is thus guaranteed \cite{shuman2012signal}. Furthermore, this improved \textbf{scalability}; learning $K$ coefficients of the predefined smooth polynomial functions meant that the number of learnable parameters was no longer dependent on the size of the input graph. Additionally, the learned model could be applied to other graphs too, as opposed to spectral filter coefficients which are basis dependant. Since then, multiple potential polynomials have been used for specialised effects (e.g. Chebyshev polynomials, Cayley polynomials \cite{levie2017gcncayley}). 

Equation~\ref{eq:eigenimp} outlines this approach. The learnable parameters are $\theta_k$ --- vectors of Chebyshev polynomial coefficients --- and $T_k(\widetilde{\matlambda})$ is the Chevyshev polynomial of order $k$ (dependent on the normalised diagonal matrix of scaled eigenvalues $\widetilde{\matlambda}$). Chevyshev polynomials can be computed recursively with a stable recurrence relation, and form an orthogonal basis \cite{tang2019chebnet}. We recommend \cite{phillips2003interpolation} for a full treatment on Chebyshev polynomials.

Interestingly, these approximate approaches demonstrate an equivalence between spatial and spectral techniques. Both are \textbf{spatially localised} and allow for a single filter to extract repeating patterns of interest throughout a graph, both have a number of learnable parameters which is independent of the input graph size, and each have meaningful and intuitive interpretations from a spatial (Figure~\ref{fig:tradconv}) and spectral (Figure~\ref{fig:gfunction}) perspective. In fact, GCNs can be viewed as a first order approximation of Chebyshev polynomials \cite{liu2020efficient}. For an in-depth treatment on the topic of GSP, we recommend \cite{stankovic2019understandinggsp} and \cite{shuman2013emerging}. 







\begingroup
\setlength\arraycolsep{2pt}
\renewcommand*{\arraystretch}{1.1}
\begin{table}[H]
    \begin{tabular}{p{4.0cm} p{9.0cm} }
        \toprule
        Approach & Applications \\
        \midrule
        
        GSP general (spectral) \cite{stankovic2019understandinggsp} & Multi-sensor temperature sensing (as a signal processing problem).  \\
        
        ChebNet (spectral) \cite{tang2019chebnet} & Various, but particularly in contexts where the functions to be approximated are high dimensional and smooth. \\
        
        CayleyNets (spectral) \cite{levie2017gcncayley} & Community detection, MNIST, citation networks, recommender systems, and other domains where specific frequency bands are of particular interest. \\
        
        \midrule
        
        MPNNs (spatial) \cite{gilmer2017neural} & Quantum chemistry, specifically molecular property prediction. \\
        
        GraphSAGE (spatial) \cite{hamilton2017graphsage} & Classifying academic papers, classifying Reddit posts, classifying protein functions, etc. \\
        
        GCNs (spatial) \cite{kipf2017} & Semi-supervised vertex classification on citation networks and knowledge graphs.  \\
        
        Residual Gated Graph ConvNets (spatial) \cite{bresson2017residual} &  Subgraph matching and graph clustering. \\
        
        Graph Isomorphism Networks (GINs) \cite{xu2018howpowerful} & Various, including bioinformatics and social network datasets. \\
        
        \midrule 
        
        CGNN benchmarking \cite{dwivedi2020benchmarking} & Extensive, including ZINC \cite{irwin2012zinc}, MNIST \cite{lecun1998mnist2}, CIFAR10 \cite{krizhevsky2009cifar}, etc. \\
        
        \midrule 
        
        GATs \cite{velivckovic2017graph} & Citation networks, protein-protein interaction. \\
        
        GATs \cite{sarlin2020superglue} & Robust pointwise correspondence of local image features. \\
        
        Gated Attention Modules \cite{zhang2018gaan} & Traffic speed forecasting. \\ 
        
        Edge GATs \cite{chen2021edge} & Citation networks, but generally any domain sensitive to relations / edge features. \\ 
        
        Graph Attention Tracking \cite{guo2021graph} & Visual tracking (i.e., similarity matching between a template image and a search region). \\ 
        
        Hyperbolic GATs \cite{zhang2021hyperbolic} & Hyperbolic domains, e.g. protein surfaces, biomolecular interactions, drug discovery, or statistical mechanics. \\

        Heterogeneous Attention Networks (HANs) \cite{wang2019heterogeneous} & Citation networks, IMBD (movie database networks), or any domain where vertices / edges are heterogeneous. \\ 
        
        GATs \cite{wang2019kgat} & Knowledge graphs and explainable recommender systems. \\ 
        
        \midrule
        
        Graphormers \cite{ying2021transformers} & Various, including quantum chemistry prediction. Particularly well suited to smaller scale graphs due to quadratic computation complexity of attention mechanisms. \\
        
        Graph Transformers (with spectral attention) \cite{kreuzer2021rethinking} & Various, including molecular graph analysis (i.e. \cite{irwin2012zinc} and similar). Particularly well suited to smaller scale graphs as above. \\

        \bottomrule
    \end{tabular}
\caption{A particular selection of often-cited works which use convolutional GNN techniques (such as those discussed in this section). Many of these algorithms are applicable to graph generally, and as such, the application column outlines the applications directly discussed in the cited paper. } 
\label{tab:cgnnapps}
\end{table}
\endgroup


\subsection{Advantages, Disadvantages, and Applications}

As noted, spatial techniques and spectral techniques each have their advantages and disadvantages. The most popular spatial techniques are localised, generally scalable, and easily interpretable as methods for extracting features of interest from neighborhoods within graphs. As a downside, some of the more popular approaches (i.e. GATs) require expensive computations that scale in compute complexity quadratically with the size of their inputs, making them unsuitable for large graphs. On the other hand, spectral techniques can also be localised, scalable, and physically interpreted, but in some cases require rigorous computations to calculate the graph Laplacian. In general, eigendecomposition-based techniques can't be used for graph inputs which have dynamic topologies, and are computationally prohibitive for large graphs.

\section{Graph Autoencoders}
\label{s:gaegeneral}

GAEs represent the application of GNNs (often CGNNs) to autoencoding. \edit{The goal of an AE can be summarised as follows: to project the inputs features into a new space (known as the latent space) where the projection has more desirable properties than the input representation. These properties may include:}

\edit{\begin{enumerate}
    \item The data being more separable (i.e. classifiable) in the latent space.
    \item The dimensionality of the dataset being smaller in the latent space than in the input space.
    \item The data being obfuscated for security or privacy concerns in the latent space.
\end{enumerate}}

\edit{A benefit of AEs in general is that they can often achieve this in an unsupervised manner --- i.e., they can create useful embeddings without any training data.} In their short history, GAEs have lead the way in unsupervised learning on graph-structured data and enabled greater performance on supervised tasks such as vertex classification on citation networks \cite{kipf2016variational}.





\subsection{Autoencoders in the Graph Domain}
\label{s:gae}

\begin{figure}[H]
    \centering
    
    \subfloat[The architecture for a simple traditional standard AE. AEs take a \edit{tensor input $X$}, alter the dimensionality via a learnable encoder NN, and thus convert said input into a latent embedding $Z$. From there, the AE attempts to reconstruct the original input, thus creating the reconstructed input $\hat{X}$ (this process forms the decoder). By minimising the reconstruction loss $\textnormal{L}=\|X-\hat{X}\|^{2}$, efficient latent space representations can be learned. This diagram shows an AE with a latent space representation that is smaller than the input size. \change{In practice, encoder NNs can use custom layers and connections to improve performance.}]{
        \includegraphics[width=0.95\textwidth]{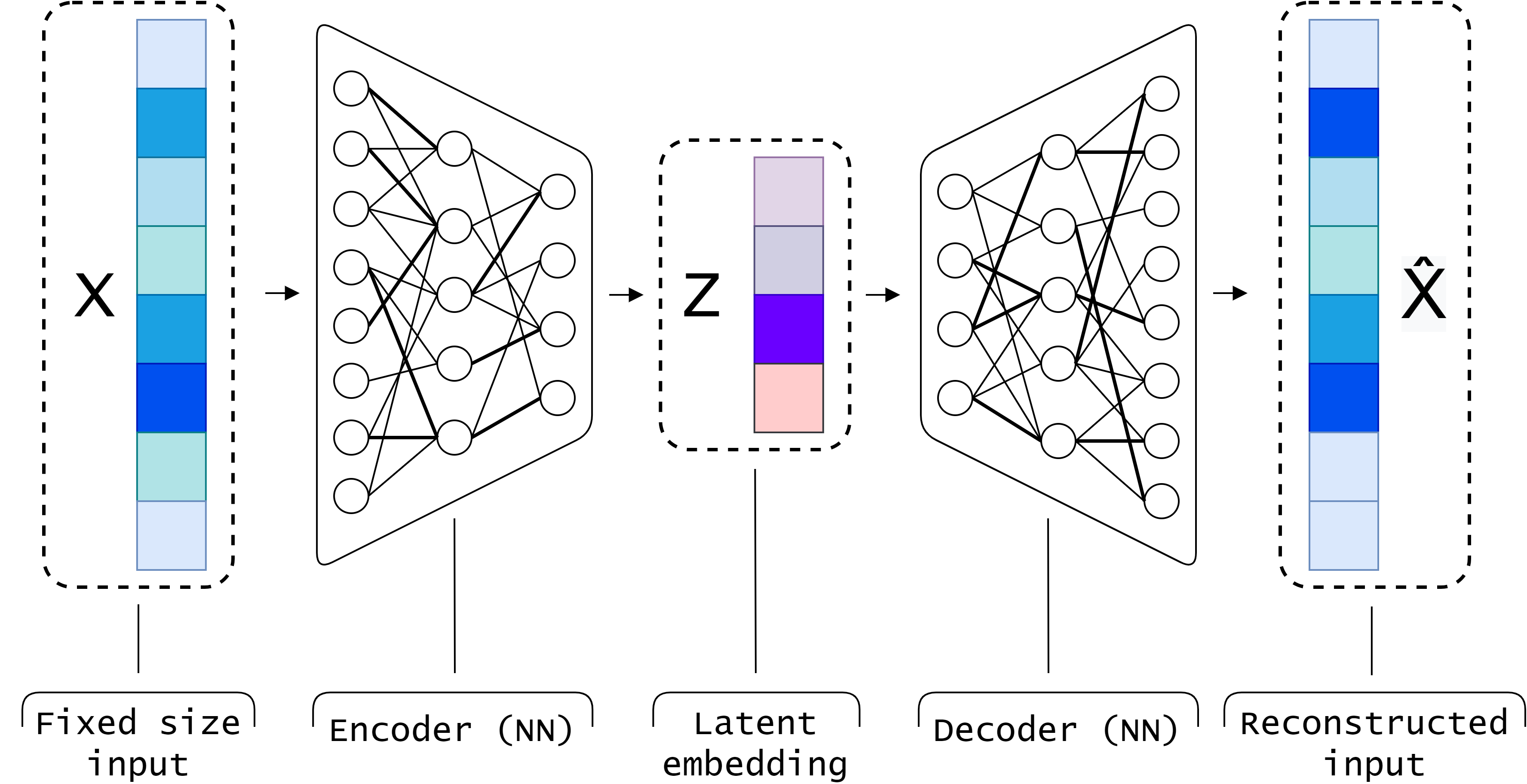}
    }
    \\
    \subfloat[\edit{The architecture for a GAE. The input graph is described by the adjacency matrix \matadjacent and the vertex feature matrix $X$ in this case (though edge and global graph features can be accepted as input also). Since the input is an unstructured graph and not a tensor, a GNN architecture, such as those described throughout this tutorial, is used to generate a matrix of latent vertex embeddings $Z$. To reconstruct the input the similarity between all pairs of latent vertex embeddings is calculated, yielding a proxy for the `connectedness' amongst the vertices in the graph. This creates the estimated adjacency matrix $\hat{A}$, which can be compared with the original \matadjacent to create a loss term. In this example, the red edges denote edges which were incorrectly reconstructed. }]{
        \includegraphics[width=0.95\textwidth]{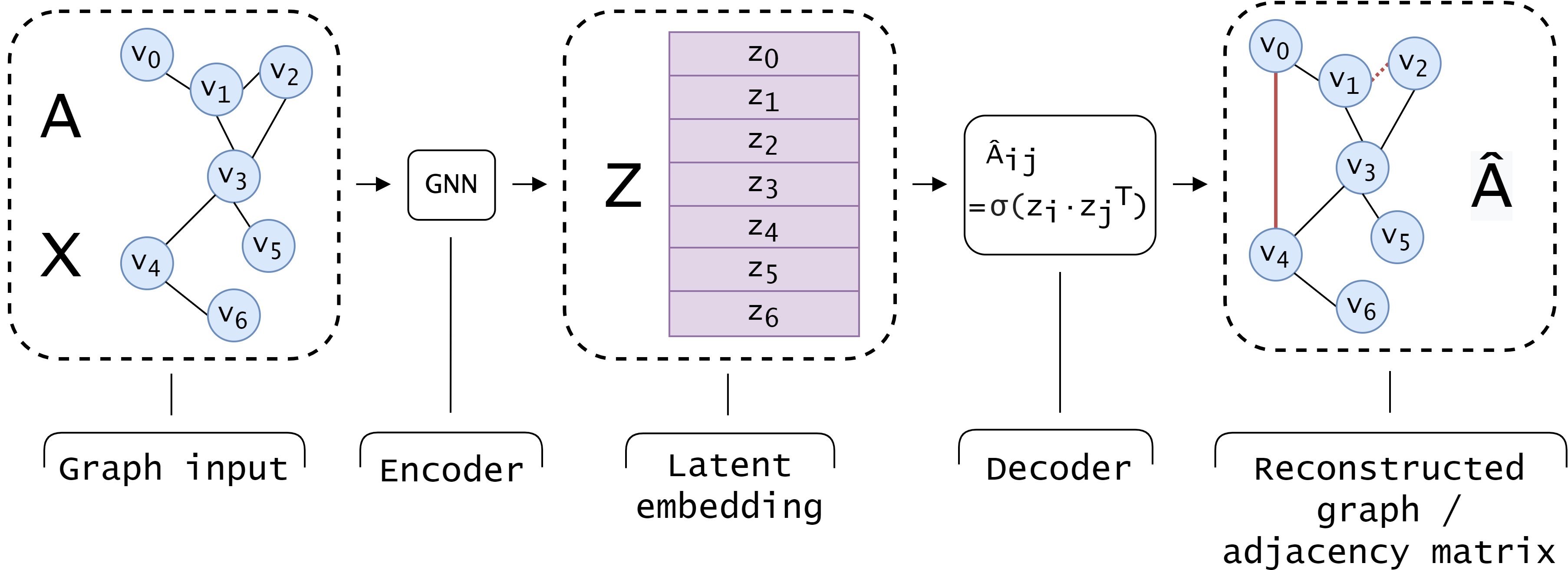}
    }
    \vspace{-2mm}
    \\
    
    \caption{The architecture for a traditional tensor-based AE, compared to a GAE.}

    \label{fig:aegae}
\end{figure}

AEs work in a two-step fashion, first encoding the input data into a latent space, and then decoding this compressed representation to reconstruct the original input data, as depicted in Figure~\ref{fig:aegae} (though in some cases higher dimensionality latent space representations have been used). The AE is then trained to minimise the \textit{reconstruction loss}, which is calculated using only the input data, and can therefore be trained in an unsupervised manner. \edit{In its simplest form, such a loss is defined as $\textnormal{Loss}_{AE}=\|X-\hat{X}\|^{2}$, where we have an input instance $X$ and the reconstructed input $\hat{X}$.}

\edit{The difference between AEs to GAEs is illustrated in Figure~\ref{fig:aegae}, and requires the definition of encoders and decoders which take in and put out graph structures respectively. One of the most common methods for doing this is to replace the encoder with a CGNN, and replace the decoder with a method that can reconstruct the graph structure of the input \cite{kipf2016variational}. }

With a well defined loss function, we can perform end-to-end learning across this network to optimize the encoding and decoding in order to strike a balance between both \textbf{sensitivity} to inputs and \textbf{generalisability} --- we do not want the network to overfit and `memorise' all training inputs. \edit{Rather, the goal is for the encoder network to represent repeating patterns within the input data in a more compressed and efficient format.} 


\edit{Once trained, GAEs (like AEs), can be split into their component networks to perform specific tasks. A popular use case for the encoder is to generating robust embeddings for supervised downstream tasks (e.g. classification, visualisation, regression, clustering, etc.), and a use for the decoder is to generate new graph instances that include properties from the original dataset. This allows the generation of large synthetic datasets.}

\subsection{Variational Graph Autoencoders}
\label{s:vgae}

\begin{figure}[H]
    \centering
    \includegraphics[width=0.95\textwidth]{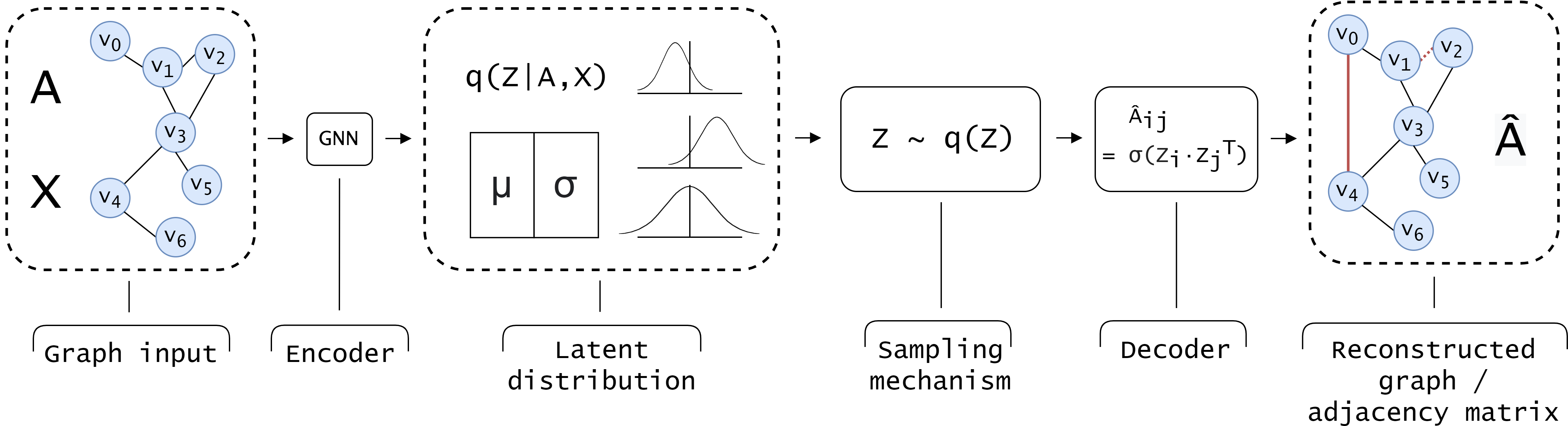}
    
    \caption{\edit{An example of a VGAE. Graph inputs are encoded via a GNN into multivariate Guassian parameters (i.e. mean and variance). These represent ranges of possible values in the latent space, which enforces a continuous latent space representation. Samples are selected from these distributions and fed to the decoder, as in Figure~\ref{fig:aegae} (b). In practice, researchers have observed that this method ensures that all regions of latent space map to meaningful outputs, and that latent vectors which are close to one another map to reconstructions that are `close to' one another in the input space. To ensure that the encoded distributions are well behaved, a penalty term is added to the loss function to enforce the distributions to match some known prior distribution (i.e., normal distributions). The total loss function for a VGAE is thus defined as $\textnormal{L}=\|X-\hat{X}\|^{2} + KL((N(0,1), q(Z))$, }}

    \label{fig:vgae}
\end{figure}

Rather than representing inputs with single points in latent space, variational autoencoders (VAEs) learn to encode inputs as probability distributions in latent space. Figure~\ref{fig:vgae} shows a VGAE which predicts a multivariate Guassian-like distribution $q(Z|A,X)$ for a given input.

\newpage
\begin{mdframed}[style=examplebox]
\vspace{-0.3cm}

\subsection*{Using Variational Graph Autoencoders for Unsupervised Learning}

\label{ss:ex4}
\small
In this example, we will implement GAEs and VGAEs to perform unsupervised learning. After training, the learned embeddings can be used for both vertex classification and edge prediction tasks, even though the model was not trained to perform these tasks initially, thus demonstrating that the embeddings are meaningful representations of vertices. We will focus on edge prediction in citation networks, though these models can be easily applied to many task contexts and problem domains (e.g., vertex and graph classification).

\begin{wrapfigure}{r}{5cm}
    \vspace{-0.8cm}
    \fbox{\includegraphics[width=4.7cm]{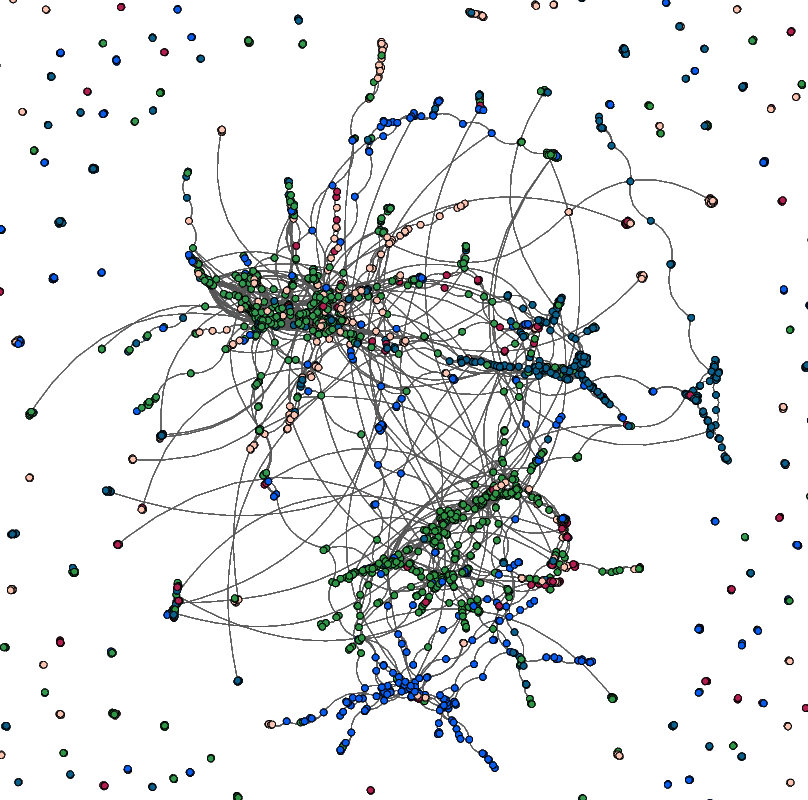}}\vspace{1mm}
    \hspace{-2mm}\fbox{\includegraphics[width=4.7cm]{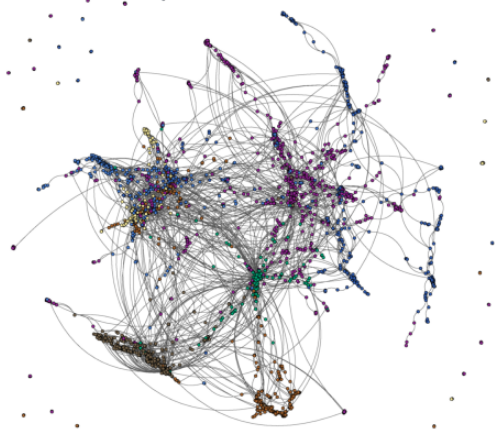}}
    \vspace{-6mm}
    \caption{Renderings of the Citeseer dataset (top) and Cora dataset (bottom). Image best viewed in colour.}
    \label{fig:Cora}
\end{wrapfigure}

\vspace{-0.2cm}
\subsection*{Dataset}
\vspace{-0.1cm}

To investigate GAEs, we use the Citeseer, Cora and PubMed datasets, which are accessible via \href{https://pytorch-geometric.readthedocs.io/en/latest/modules/datasets.html}{PyTorch Geometric} \cite{fey2019pytorch}. In each of these graphs, the vertex features are word vectors indicating the presence or absence of predefined keywords (see Section~\ref{ss:ex2} for another example of the Cora dataset being used). 

\vspace{-0.2cm}
\subsection*{Algorithms}
\vspace{-0.1cm}

We first implement a GAE. This model uses a single GCN to encode input features into a latent space. An inner product decoder is then applied to reconstruct the input from the latent space embedding (as described in Section~\ref{s:vgae}). During training we optimise the network by reducing the reconstruction loss. We then apply the model to an edge prediction task to test whether the embeddings can be used in performing downstream machine learning tasks, and not just in reconstructing the inputs. 

We then implement the VGAE as first described in \cite{kipf2016variational}. Unlike GAEs, there are now two GCNs in the encoder model (one each for the mean and variance of a probability distribution). The loss is also changed to Kullback–Leibler (KL) divergence in order to optimise for an accurate probability distribution. From here, we follow the same steps as for the GAE method: an inner product decoder is applied to the embeddings to perform input reconstruction. Again, we will test the efficacy of these learned embeddings on downstream machine learning tasks. 

\vspace{-0.2cm}
\subsection*{Results and Discussion}
\vspace{-0.1cm}
To test GAEs and VGAEs on each graph, we average the results from $10$ experiments. In each experiment, the model is trained for $200$ iterations to learn a 16-dimensional embedding.


\begingroup
\setlength{\intextsep}{0pt}%
\setlength{\columnsep}{0pt}%
\begin{wraptable}{l}{8cm}
    \vspace{-0.4cm}
    \begin{tabular}{|l|l|c|c|}
        \hline
        Algorithm & Dataset & AUC  & AP$^{a}$ \\
        \hline
        GAE & Citeseer & 0.858 ($\pm$0.016) & 0.868 ($\pm$0.009) \\
        \textbf{VGAE} & Citeseer & \textbf{0.869 ($\pm$0.007)} & \textbf{0.878 ($\pm$0.006)} \\
        GAE & Cora & 0.871 ($\pm$0.018) & 0.890 ($\pm$0.013) \\
        \textbf{VGAE} & Cora & \textbf{0.873 ($\pm$0.01)} & \textbf{0.892 ($\pm$0.008)} \\
        \textbf{GAE} & PubMed & \textbf{0.969 ($\pm$0.002)} & \textbf{0.971 ($\pm$0.002)} \\
        VGAE & PubMed & 0.967 ($\pm$0.003) & 0.696 ($\pm$0.003) \\
        \hline
    \end{tabular}
    \caption{Comparing the link prediction performance of autoencoder models on Citeseer.}
    \label{tab:autoencres}
\end{wraptable}
\endgroup

\noindent In alignment with \cite{kipf2016variational}, we see the VGAE outperform GAE on the Cora and Citeseer graphs, while the GAE outperforms the VGAE on the PubMed graph. Performance of both algorithms was significantly higher on the PubMed graph, likely owing to PubMed's larger number of vertices ($\numV = 19717$), and therefore more training examples, than Citeseer ($\numV = 3327$) or Cora ($\numV = 2708$). Moreover, while Cora and Citeseer vertex features are simple binary word vectors (of sizes $1433$ and $3703$ respectively), PubMed uses the more descriptive TF-IDF word vector, which accounts for the frequency of terms. This feature may be more discriminative, and thus more useful when learning vertex embeddings.


\vspace{0mm}  

\end{mdframed}

This creates a `smoother' latent space that covers the full spectrum of inputs, rather than leaving `gaps', where an unseen latent space vector would be decoded into a meaningless output. This has the effect of increasing generalisation to unseen inputs and regularising the model to avoid overfitting. Ultimately, this approach transforms the GAE into a more suitable generative model.


Unlike in GAEs --- where the loss is simple the mean squared error between the input and the reconstructed input --- a VGAE's loss \edit{imposes an additional penalty which ensures that the latent distributions are normalised.}  \edit{More specifically, this term regularises the latent space distributions by ensuring that they do not diverge significantly from some \edit{prior} distribution with desirable properties. In our example, we use the normal distribution (denoted as $N(0, 1)$). This divergence is quantified in our case using \textbf{Kulback-Leibler divergence} (denoted as `KL'), though other similarity metrics (e.g. Wassertein space distance or ranking loss) can be used successfully. Without this loss penalty, the VGAE encoder might generate distributions with small variances, or high magnitude means: both of which would make it harder to sample from the distribution effectively.}

\subsection{\edit{Improving Robustness with Graph Adversarial Techniques}}
\label{s:gadvt}

Graph Adversarial Techniques (GAdvTs) use adversarial learning methods whereby an AI model acts as an adversary to another during training to mutually improve the performance of both models in tandem. Due to the adversarial nature of GAdvT's, developments in this area have been described as an ``arms race between attackers and defenders'' \cite{chen2020survey}. As with traditional adversarial techniques, common goals for GAdvTs include:

\begin{itemize}
    \item Improving the robustness, regularisation, or distribution of learned embeddings.
    \item Improving the robustness of models to targeted attacks.
    \item Training generative AI models.
\end{itemize}



\begin{figure}[H]
    \centering
    \includegraphics[width=0.95\textwidth]{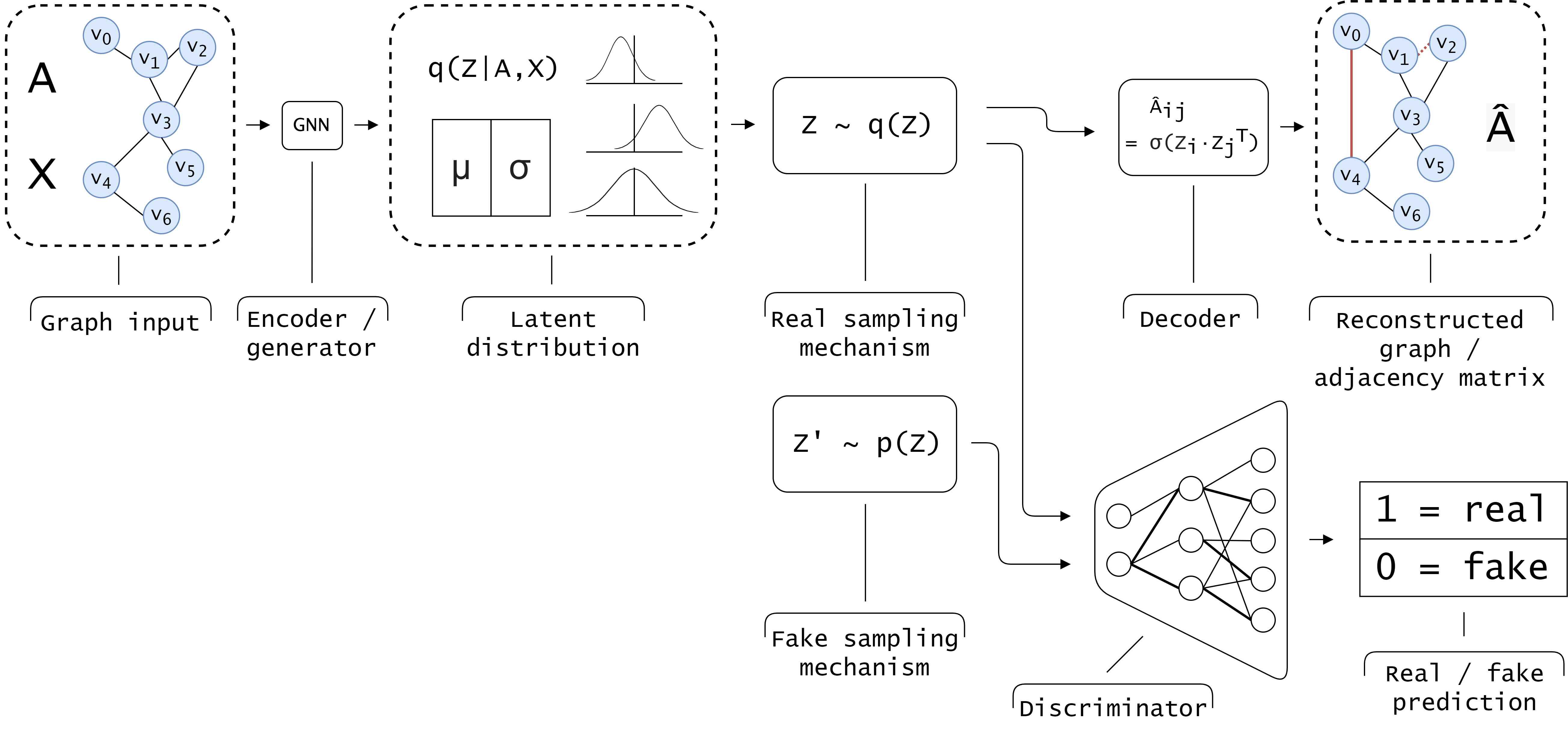}
    
    \caption{A typical approach to adversarial training with VGAEs. The top row described a VGAE as illustrated in Figure~\ref{fig:vgae}. Importantly, for each real sample, a `fake' sample is generated from some prior distribution $p(Z)$ (e.g., a multivariate Guassian, or some other distribution which is believed to model the properties of the latent space attributes). During training, these fake and real samples are input into a discriminator network, which predicts whether said inputs are real or fake. If the discriminator \textit{correctly} classifies the sample, the generator is \textit{penalised}, thus optimising the encoder to generate distributions whose samples are more likely to `fool' the discriminator. In other words, this causes the encoder to create samples which have similar properties to the samples pulled from the prior distribution $p(z)$, thus acting as a form of regularisation.     }

    \label{fig:adversary}
\end{figure}

The field of GAdvT's is broad, with multiple different kinds of attacks, training regimes, and use cases. In this tutorial, we'll look at how a GAdvTs can be used to extend VGAEs to create robust encoding networks, and well regularised generative mechanisms. Figure~\ref{fig:adversary} describes a typical architecture for adversarially training a VGAE. 

To ensure that the sampling operation is differentiable, VGAEs levereage a `reparameterisation trick', where a random Guassian sample is generated from $N(0,1)$ \textit{outside} the forward pass of the network. The sample is then transformed by the parameterisation of the generated distribution $q(z)$, rather than having the sample be generated directly from $q(z)$ \cite{doersch2016tutorial}. Since this approach is entirely differentiable, it allows for end-to-end training via backpropagation of an unsupervised loss signal.

\subsection{Advantages, Disadvantages, and Applications}

In this section, we have explained the mechanics behind traditional and graph autoencoders. Analagous to how AEs use NN to perform encoding, GAEs use CGNNs to perform encoding and create embeddings \cite{kipf2016variational}. Similarly, an unsupervised reconstruction error term is defined; in the case of GAEs, this is between the original adjacency matrix and the predicted adjacency matrix (produced by the decoder). GAEs and VGAEs represent a simple method for performing unsupervised training, which allows us to learn powerful embeddings in the graph domain without any labelled data, \edit{but requires regularisation techniques to smooth their latent space representations, and reparameterisation tricks to ensure differentiability}. 

Before the seminal work in \cite{kipf2016variational} on VGAEs, a number of deep GAEs had been developed for unsupervised training on graph-structured data, including Deep Neural Graph Representations (DNGR) \cite{cao2018graph} and Structure Deep Network Embeddings (SDNE) \cite{wang2016structural}. These methods operate on only the adjacency matrix, so information about both the entire graph and the local neighbourhoods is lost. More recent work mitigates this by using an encoder that aggregates information from a vertex's local neighbourhood to learn latent vector representations. For example, \cite{salha2019simple} proposes a linear encoder that uses a single weights matrix to aggregate information from each vertex's one-step local neighbourhood, showing competitive performance on numerous benchmarks. Despite this, typical GAEs use more complex encoders --- primarily CGNNs --- in order to capture nonlinear relationships in the input data and larger local neighbourhoods \cite{kipf2016variational, cao2018graph, wang2016structural, yu2018netRA, berg2017graph, tu2018recursive, Bojchevski2018graph2gauss, pan2018arga}. 


\vspace{-2mm}

\begingroup
\setlength\arraycolsep{2pt}
\renewcommand*{\arraystretch}{1.1}
\begin{table}[H]
    \begin{tabular}{p{4.0cm} p{9.0cm} }
        \toprule
        Approach & Applications \\
        \midrule
        
        Deep Neural Graph Representation \cite{cao2018graph} & Various, including clustering, calculating useful vertex embeddings, and visualisation. \\ 
        
        Structure Deep Network Embeddings \cite{wang2016structural} & Various, including language networks, citation networks, and social networks. \\
        
        Denoising Attribute AEs \cite{hettige2020rase} & Various, including social networks and citation networks. \\ 
        
        
        Link prediction-based GAEs (and VGAEs) \cite{salha2019directed} & Various, including link prediction and bidirectionally prediction on citation networks. \\
        
        \midrule
        
        VGAEs \cite{kipf2016variational} & Various, including citation networks. \\
        
        Deep Guassian Embedding of Graphs (G2G) \cite{Bojchevski2018graph2gauss} & Various, including citation networks. \\
        
        Semi-implicit VGAEs \cite{hasanzadeh2019semiimplicit} & Various graph analytic tasks, including citation networks.  \\
        
        Adversarially Regularised AEs (NetRA) \cite{yu2018netRA} & Directed communication networks, software class dependency networks, undirected social networks, citation networks, directed word networks with inferred `Part-of-Speech' tags, and Protein-Protein Interactions. \\
        
        Adversarially Regularised Graph Autoencoder (ARGA, and its variants) \cite{pan2018arga} & Various, including vertex clustering and visualisation of citation networks. \\
        
        
        
        
        \midrule
        
        Graph Convoltuional Generative Adversarial Networks \cite{vinchoff2020gantraffic} & Traffic prediction in optical networks (particularly in domains with `burst events'). \\ 
        
        FeederGAN (adversarial) \cite{liang2020feedergan} & Generation of distributed feeder circuits. \\ 
        
        Labelled Graph GANs \cite{fan2019labeledgan} & Generating graph-structured data with vertex labels. Demonstrated for citation networks and protein graphs. \\ 
        
        Graph GANs for Sparse Data \cite{kansal2020ganphysics} & Generating sparse graph datasets. Demonstrated for MNIST and high energy physics proton-proton jet particle data. \\ 
        
        
        Graph Convolutional Adversarial Networks \cite{hong2019ganlongitudinal} & Predicting missing infant diffusion MRI data for longitudinal studies. \\

        \bottomrule
    \end{tabular}
\caption{A selection of works using GAE / VGAE / GAdvT techniques as discussed in this section. } 
\label{tab:gaeapps}
\end{table}
\endgroup

\vspace{-15mm}

\edit{\section{Future Research}}
\label{s:future}

The field of GNNs is rapidly developing, and there are numerous directions for meaningful future research. In this section, we outline a few specific directions which have been identified as important research areas to focus on \cite{zhou2018methapps, zhang2018survey, wu2019comprehensive, yuan2020explainability}.

\subsection{Explainability}

Recent advancements in deep learning have allowed deeper NNs to be developed and applied throughout the field of AI. As the mechanics that drive predictions (and thus decisions) become more complex, the path by which those decisions are reached becomes more obfuscated. Explainable AI (XAI) promises to address this issue.












\newpage Explainability in the graph domain promises much of the same benefits as it does across AI, including more interpretable outputs, clearer relationships between inputs and outputs, more interpretable models, and in general, more trust between AI and human operators across problem domains (e.g. digital pathology \cite{jaume2020towards}, knowledge graphs \cite{wang2019explainable}, etc.). 

While the suite of available XAI algorithms has been consistently growing over the recent years (e.g. LIME, SHAP), graph specific XAI algorithms are relatively few and far between \cite{yuan2020explainability}. A key reason for this might be the requirement for graph explanations to incorporate not just the relationships among the input features, but also the relationships surrounding the input's structural / topological information. In particular, the exploration of instance-level explainers --- including high fidelity perturbative and gradient-based methods --- may provide good approximations of input importance in graph prediction tasks. Further techniques, especially those which assign quantitative importances to a graph's structure \textit{and} its features will give a more holistic view of explainability in the graph domain. 



\subsection{Scalability}

In traditional deep learning, a common technique for dealing with extremely large datasets and AI models is to distribute and parallelise computations where possible. In the graph domain, the non-rigid structure of graphs presents additional challenges. For example; how can a graph be uniformly partitioned across multiple devices? How can message passing frameworks be efficiently implemented in a distributed system? These questions are especially pertinent for extremely large graphs. Recent developments suggest that sampling based approaches may provide appropriate solutions in the near future \cite{zheng2020distdgl}, though such solutions are non-trivial, especially when graphs are stored on distributed systems \cite{serafini2021scalable}.

Moreover, the scalability of GNN modules themselves may be improved by further directed research. For example, popular GNN variants such as MPNNs can in practice only be applied to small graphs due to the large computational overheads associated with the message passing framework. Methods such as GATs show promising results regarding scalability, but attentional mechanisms still incur a quadratic time complexity, which may be prohibitive for graphs with large neighborhoods (on average). An exciting further avenue of research regarding GATs is their equivalence to Transformer networks \cite{ying2021transformers, kreuzer2021rethinking, vaswani2017transformer, joshi2020transformers}. Further directed research in this area may contribute not only to the development of exciting new graph-based techniques, but also the understanding of Transformer networks as a whole. Breakthroughs in this area may address challenges specific to Transformers, such as the design of efficient positional encodings, effective warm-up strategies, and the quantification of inductive biases.  







\subsection{Advanced Learning Paradigms}

Self-supervised Learning (SSL) techniques have recently been suggested as the `next step' in AI training paradigms, as they close the gap --- and even outperform --- fully supervised approaches in visual tasks \cite{caron2021emerging, assran2021semi}. Contemporary approaches to SSL include using models that learn from one another \cite{grill2020bootstrap, caron2021emerging, zbontar2021barlow}. In related work, recent research suggests that contrastive objectives can be designed in the graph domain by selecting views of a single graph instance, thus permitting the capture of universal structural properties across graph without the need for large labelled datasets \cite{zhu2021graph, qiu2020gcc, jovanovic2021towards, zhu2020deep}. These initial investigations demonstrate that the graph domain is well suited to the application of advanced learning paradigms techniques. Further research in this area may produce more general pretrained GNNs, allow the leveraging of large unlabelled graph datasets, and yielding further insight into nature unsupervised / weakly supervised learning in the development of intelligence.








\section{Conclusion}
\label{s:conc}

The development of GNNs has accelerated \textit{hugely} in the recent years due to increased interest in exploring unstructured data and developing general AI solutions. \edit{In this paper, we have illustrated key GNN variants, described the mechanisms which underpin their operations, addressed their limitations, and worked through examples of their application to various real world problems (with links to more advanced literature where necessary). Going forward, we expect that GNNs will continue to emerge as an exciting and highly performant branch of algorithms that natively model and address important real-world problems}. 

\section*{Funding}

This work was partially supported by ISOLABS, the Australian Research Council (Grants DP150100294 and DP150104251), the National Natural Science Foundation of China (No. U20A20185, 61972435), the Natural Science Foundation of Guangdong Province (2019A1515011271), and the Shenzhen Science and Technology Program (No. RCYX20200714114641140, JCYJ20190807152209394).

\begin{acks}

We express a special appreciation to Josh Crowe at ISOLABS for his ongoing support of technical research (including this tutorial paper) at ISOLABS. We also thank Richard Pienaar for providing early feedback which greatly improved this work.

\end{acks}



\bibliographystyle{ACM-Reference-Format}
\bibliography{main}

\appendix
\label{s:appendix}

\end{document}